\documentclass[11pt]{article}

\usepackage[preprint]{acl}

\usepackage{times}
\usepackage{latexsym}
\usepackage{enumitem}
\usepackage{bbding}
\usepackage{svg}

\usepackage[T1]{fontenc}
\usepackage[utf8]{inputenc}

\usepackage{microtype}

\usepackage{inconsolata}
\usepackage{amssymb}
\usepackage{subcaption}
\usepackage{booktabs}
\usepackage{xcolor}
\usepackage{multirow}   %用于多行合并（虽然此表主要用多列）
\usepackage{colortbl}   %用于表格行背景色
\usepackage[titles]{tocloft}   % 可选，让目录样式更美观
\usepackage{etoc}          % ① 用 etoc 控制目录范围
\definecolor{lightgraybg}{gray}{0.90} % 浅灰色背景，用于分隔不同设置
\definecolor{lightgreenbg}{RGB}{230, 255, 230}   % 可自行微调 RGB
\definecolor{lightredbg}{RGB}{252, 245, 225} % 浅黄色背景，用于分隔不同设置
\definecolor{navyblue}{RGB}{0, 80, 180} % 深蓝色，用于高亮最优结果
\definecolor{textgray}{gray}{0.5}     % 灰色文字，用于 Upper Bound
\definecolor{textred}{RGB}{215, 65, 50}
\definecolor{textgreen}{RGB}{0,102,51}
\usepackage{indentfirst}
\usepackage{amsmath}
\DeclareMathOperator*{\argmax}{arg\,max}
\usepackage{svg} 
\usepackage{multirow}
\usepackage{amsthm}
\usepackage{subcaption}

\newtheoremstyle{mydef}
  {3pt}   % 上方间距
  {3pt}   % 下方间距
  {\normalfont} % 正文字体
  {}      % 缩进
  {\bfseries} % 标题字体
  {.}     % 标题后标点
  {0.5em} % 标题后空格
  {\thmname{#1}\ \thmnumber{#2}\ \thmnote{\textit{(#3)}}}

\theoremstyle{mydef} % 正文字体为正常直立体

\usepackage{graphicx}
\usepackage[most]{tcolorbox} % 画彩色盒子
\newtcolorbox{keytakeaway}{
  enhanced,
  colback=green!1!white,     % 背景色（很浅的蓝色）
  colframe=green!20!black,   % 边框颜色（深蓝）
  boxrule=1.2pt,            % 边框线宽
  arc=3mm,                  % 圆角程度
  left=2mm, right=2mm,      % 内边距
  top=1mm,  bottom=1mm
}
\title{From Inheritance to Saturation: Disentangling the Evolution of Visual Redundancy for Architecture-Aware MLLM Inference Acceleration}
\author{Jiaqi Shi$^1$ \and Yuechan Li$^2$ \and Xulong Zhang$^3$ \and Xiaoyang Qu$^3$ \and Jianzong Wang$^3$\\
        $^1$University of Science and Technology of China \\
        $^2$Wuhan University \\
        $^3$Ping An Technology (Shenzhen) Co., Ltd., Shenzhen, China}

\begin{document}

% \etocstoplocaltoc           % 关闭本地目录
% \etocsettocdepth{none}      % 不写任何条目到 .toc 文件
\maketitle
\begin{abstract}
High-resolution Multimodal Large Language Models (MLLMs) face prohibitive computational costs during inference due to the explosion of visual tokens. Existing acceleration strategies, such as token pruning or layer sparsity, suffer from severe "backbone dependency", performing well on Vicuna or Mistral architectures (e.g., LLaVA) but causing significant performance degradation when transferred to architectures like Qwen. To address this, we leverage truncated matrix entropy to uncover a universal three-stage inference lifecycle, decoupling visual redundancy into universal Intrinsic Visual Redundancy (IVR) and architecture-dependent Secondary Saturation Redundancy (SSR). Guided by this insight, we propose HalfV, a framework that first mitigates IVR via a unified pruning strategy and then adaptively handles SSR based on its specific manifestation. Experiments demonstrate that HalfV achieves superior efficiency-performance trade-offs across diverse backbones. Notably, on Qwen25-VL, it retains 96.8\% performance at a 4.1$\times$ FLOPs speedup, significantly outperforming state-of-the-art baselines.  Our code is available at \url{https://github.com/civilizwa/HalfV}.
\end{abstract}

\section{Introduction}
% Vision--Language Models (VLMs) have recently made rapid progress\citep{llava,internvl,qwen2vl,llava-vl,qwen25vl}. 
% Despite their capability, VLMs suffer from substantial inference costs. 
% Under high-resolution inputs, the sequence is often dominated by visual tokens produced by a Vision Transformer (ViT) encoder \citep{vit}. 
% Given the $\mathcal{O}(N^2)$ complexity of self-attention, this token imbalance becomes a key bottleneck for latency-sensitive and resource-constrained deployment.
Multimodal Large Language Models (MLLMs) have witnessed remarkable advancements \citep{llava,internvl,qwen2vl,llava-vl,qwen25vl}. However, their practical deployment is severely impeded by prohibitive computational costs. Specifically, under high-resolution settings, the input sequence becomes overwhelmingly dominated by visual tokens encoded by the Vision Transformer (ViT) \citep{vit}. Given the quadratic $\mathcal{O}(N^2)$ complexity of self-attention, this visual token explosion incurs a massive computational burden primarily during the prefill stage, thereby creating a critical bottleneck for latency-sensitive applications.
\begin{figure}[t]
    \centering
    \includegraphics[width=\linewidth]{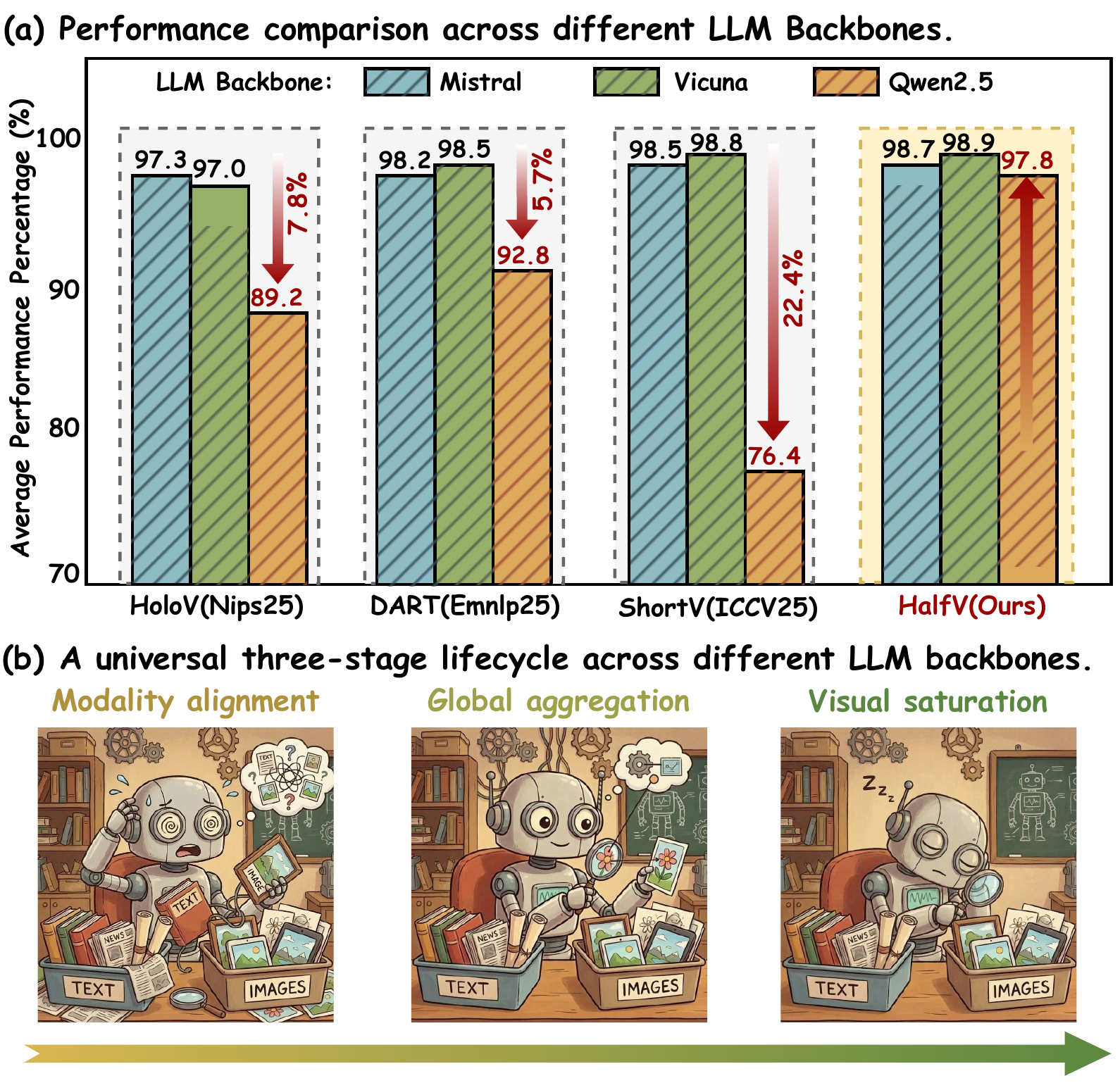}
    \caption{(a): Comparison of token-level methods (HoloV, DART) at a 77.8\% pruning ratio and the layer-level method (ShortV) across Vicuna, Mistral and Qwen backbones. Results represent the average relative performance compared to the baseline (\%) across POPE, MME, MMBench, and SQA datasets; (b): We find that models with different backbones all exhibit a universal three-stage lifecycle: Modality alignment, global aggregation and visual saturation.}
    %(a):目前主流的视觉Token裁剪方法(fastv,Holov,Dart)和层冻结方法(ShortV)在LLava-1.5v-7B和Qwen25-VL-7B中的测试结果比较，图中结果均取自POPE、MME、MMB和SQA中的结果与各个Baseline百分比的平均值 
    % 我们发现不同架构的模型均存在图中所示的三个阶段:
    \label{fig:first_pic}
    \vspace{-0.4cm}
\end{figure}

Current acceleration efforts, encompassing token-level visual token pruning \citep{fastv,llava-prumerge,sparsevlm,flowcut,fitprune,pdrop,feather} and layer-level layer sparsity \citep{shortv,dyvte}, have attempted to mitigate this cost. However, we observe that these approaches exhibit severe \textit{backbone dependency}, often being overfitted to Vicuna or Mistral architectures (e.g., LLaVA series \citep{llava-vl,llavanext}) while neglecting the heterogeneity of underlying LLM backbones. As shown in Figure \ref{fig:first_pic}, transferring these strategies to the Qwen2.5 backbone results in performance degradation ranging from 5.7\% to 22.4\%. Crucially, by utilizing LLaVA-Next \citep{llavanext} which employs dynamic resolution as a control variable, we rule out the possibility that this failure stems from the visual front-end (More details see Appendix \ref{appendix:tme}). While LLaVA-Next remains robust to existing pruning methods, Qwen2.5-VL exhibits distinct sensitivity. \textbf{This firmly establishes that the bottleneck lies in the intrinsic mechanism by which different LLM backbones process visual information.}

To decode this mechanism and break the architecture barrier, we employ \textbf{\textit{truncated matrix entropy}} \citep{truncated,uncomp} as a probe to systematically trace the evolution of visual information. Our analysis uncovers a universal three-stage lifecycle across architectures: I. Modality Alignment, II. Global Aggregation and III. Visual Saturation. Based on this evolution, we categorize the observed redundancies into two distinct types: 

\textbf{(1) Intrinsic Visual Redundancy (IVR)}: Dominating Stage I, this redundancy stems from ViT's dense tokenization, where highly correlated, spatially adjacent patches are mapped into the LLM space with minimal interaction. 

\textbf{(2) Secondary Saturation Redundancy (SSR)}: Emerging in Visual Saturation (Stage III), this redundancy is a direct byproduct of Global Aggregation (Stage II). After the LLM aggregates dispersed visual evidence into key semantic regions, the deep layers reach a state of semantic saturation where additional computation yields diminishing information gain. 
Notably, while IVR is universal, the physical manifestation of SSR is architecture-dependent: it appears as layer-level inactivity in Vicuna/Mistral backbones but as extreme token sparsity in Qwen backbones.

Guided by the distinct mechanisms, we propose HalfV, an architecture-aware acceleration framework that decouples redundancy reduction into two steps. Specifically, HalfV targets the universal IVR with a unified pruning strategy applicable across models, while addressing the architecture-dependent SSR with an adaptive reduction mechanism that tailors the acceleration paradigm to each backbone's unique saturation manifestation. By disentangling this complex evolution process, HalfV achieves a superior efficiency-performance frontier. Extensive experiments across diverse architectures validate the effectiveness of our method, providing a principled perspective for designing future universal MLLM acceleration.

Our contributions are summarized as follows:
\begin{itemize}[leftmargin=*, noitemsep, topsep=0pt]
    \item \textbf{Unveiling Backbone Heterogeneity}: We conduct a systematic evaluation of existing acceleration strategies and identify a critical "backbone dependency". We find that this dependency stems from intrinsic LLM processing mechanisms rather than visual front-end differences.
    \item \textbf{Decoupling Redundancy Mechanisms}: Utilizing an entropy-based probe, we decouple visual redundancy into universal Intrinsic Visual Redundancy (IVR) and architecture-dependent Secondary Saturation Redundancy (SSR). 
    \item \textbf{HalfV Framework}: We propose HalfV, an architecture-aware framework that aligns with this redundancy evolution. It employs a unified pruning strategy for the universal IVR and an adaptive reduction mechanism targeting the specific SSR manifestation of each backbone, achieving a superior efficiency-performance frontier across diverse architectures.
\end{itemize}

\section{Related Work}
\begin{figure*}[htbp]
    \centering
    % 定义基准高度 (根据需要微调，例如 4.5cm)
    \newlength{\subfigheight}
    \setlength{\subfigheight}{3.5cm} 

    % --- 第一区域：图片 A (左) ---
    % 减小宽度：从 0.35 改为 0.31 (根据图片实际长宽比微调)
    % --- 第二区域：图片 B (中) ---
    \begin{minipage}[b]{0.24\textwidth}
        \centering
        \begin{subfigure}{\linewidth}
            \includegraphics[height=\subfigheight]
            % {pics/spectrum/spectrum_qwen_gqa_100.png}
            {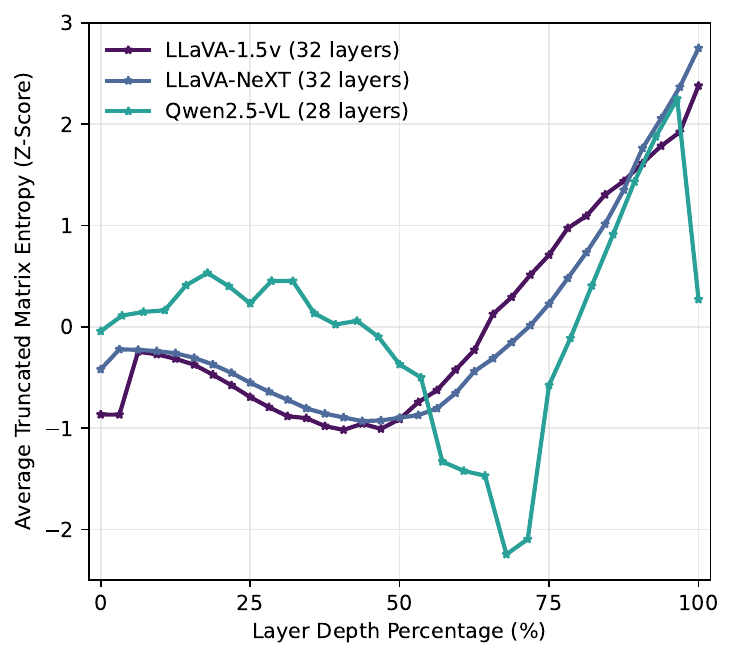}
            % \caption{$TME$ for Different Architectures}
            \caption{}
            \label{fig:tme_model}
        \end{subfigure}
    \end{minipage}
    % \vspace{0em}
    \hfill
    \begin{minipage}[b]{0.24\textwidth}
        \centering
        \begin{subfigure}{\linewidth}
            % 核心：高度固定，宽度自适应，但不能超过 linewidth
            \includegraphics[height=\subfigheight]
             % {pics/four_pics/Three_model_tme.pdf}
            {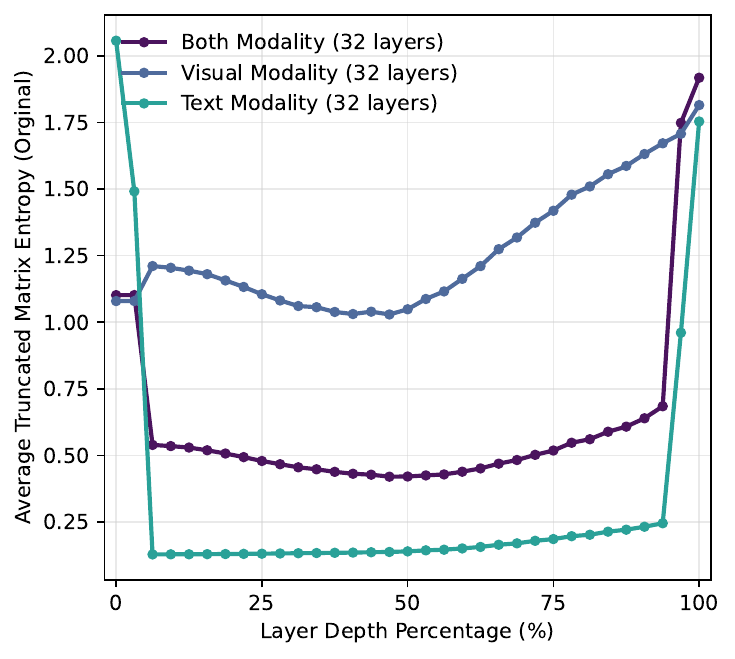}
            % \caption{$TME$ for Different Tokens on LLaVA-1.5v-7B}
            \caption{}
            \label{fig:tme_modality}
        \end{subfigure}
    \end{minipage}
    % \vspace{0em}
    \hfill
    \begin{minipage}[b]{0.25\textwidth}
        \centering
        \begin{subfigure}{\linewidth}
            \includegraphics[height=\subfigheight]
            % {pics/four_pics/three_modality_tme_llava_7b.pdf}
            {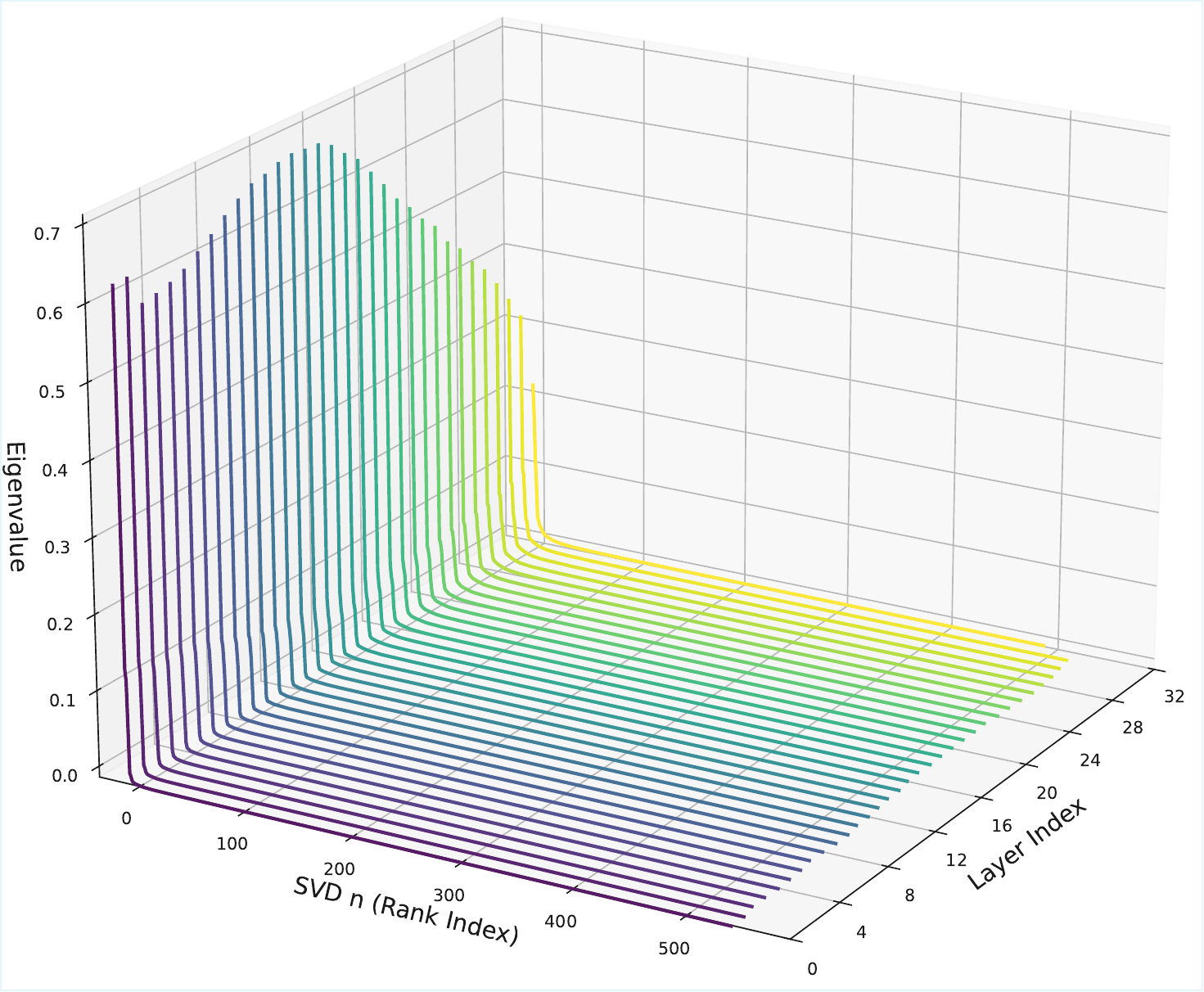}
            % \caption{3D Line Plot of Spectrum}
            \caption{}
            \label{fig:3dplot}
        \end{subfigure}
    \end{minipage}
    % \vspace{0em}
    \hfill
    \begin{minipage}[b]{0.24\textwidth}
        \centering
        \begin{subfigure}{\linewidth}
            \includegraphics[height=\subfigheight]
            % {pics/four_pics/redundancy.pdf}
            {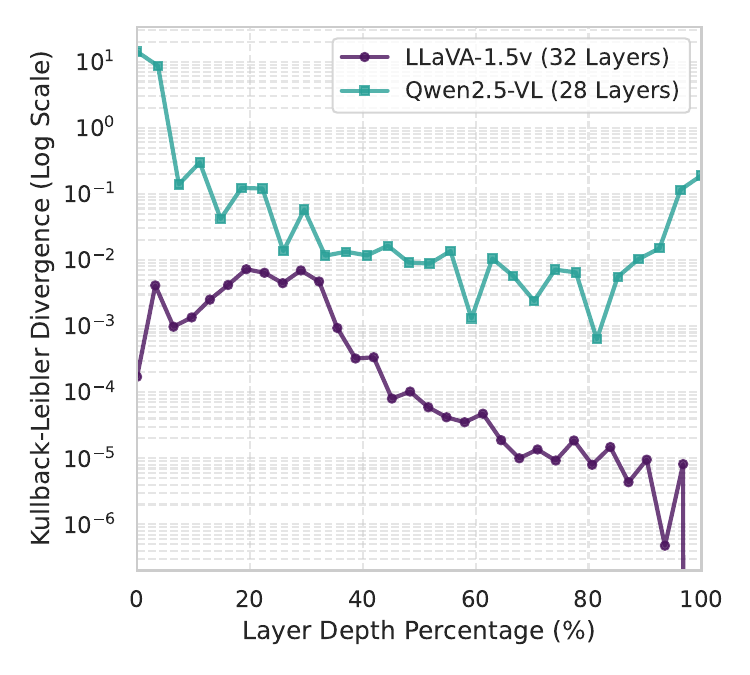}
            % \caption{Truncated Matrix Entropy}
            \caption{}
            \label{fig:kl}
        \end{subfigure}
    \end{minipage}
    \caption{(a) The average truncated matrix entropy for different architectures on GQA dataset. The x-axis is the depth percentage of the layer; (b) The average truncated matrix entropy for different types of tokens on LLaVA-1.5v-7B.; (c) 3D Line Plot of the Average Spectrum of $\mathcal{G}$ across Layers on LLaVA-1.5v-7B. Additional results see Appendix \ref{appendix:spectrum}; (d) KL divergence for Vicuna and Qwen2.5 backbones on AI2D dataset.}
    \vspace{-0.4cm}
    \label{fig:three_columns}
\end{figure*}
\subsection{Multimodal Inference Acceleration}

Our work focuses on inference acceleration during the prefill stage of MLLMs. From a redundancy perspective, existing methods fall into two categories: (i) Token-level redundancy, which retains a subset of visual tokens via heuristic scoring rules \citep{fastv,sparsevlm,feather,topv,pdrop,dart,divprune}. (ii) Layer-level redundancy. ShortV \citep{shortv} shortens the effective inference path by suppressing visual state updates in layers with low contribution to the final output. VTW \cite{vtw} assumes that visual information has been transferred to text in deeper layers, and thus removes all visual tokens in deeper layers for higher speed. 

\subsection{Internal Evolution of Representations} 
Understanding the internal evolution of representations has been widely recognized as a prerequisite for model optimization. Prior works analyzed internal signals from three perspectives: (i) intermediate representations \cite{rep}, using probes or classifiers to evaluate how layers encode objects, relations, and semantics; (ii) activation patterns and channel sparsity \cite{act}, to understand information flow; (iii) attention mechanism, with \citet{dyvte} summarizing MLLM inference into three stages and using a controller to dynamically discard visual tokens for speedup. However, it relies on heuristic stage boundaries and introduces additional training costs and complexity. In contrast, our work provides a principled foundation for stage-aware acceleration without auxiliary training or controllers.  

\section{Method}
In this section, we first introduce truncated matrix entropy and use it as a unified probe to track redundancy across inference depth. Building on these observations, we present \textbf{HalfV}, a two-step acceleration framework that aligns acceleration strategies with redundancy evolution.

\subsection{Preliminary: Truncated Matrix Entropy}
\label{sec:tme}
In each transformer layer, a set of tokens from a given modality is represented by high-dimensional hidden states. Let $\mathbf{h}_i \in \mathbb{R}^{D}$ denote the hidden states of the $i$-th token at a certain layer, where $D$ is the hidden dimension.
For a group of $N$ tokens, we stack its hidden states into a representation matrix
$\mathcal{Z} = [\mathbf{h}_1, \mathbf{h}_2, \ldots, \mathbf{h}_N]^\top \in \mathbb{R}^{N \times D}$, where $N$ is the number of tokens. The Gram matrix $\mathcal{G}$ is then computed from $\mathcal{Z}$ as:
\begin{equation}
\small
\label{gram}
    \mathcal{G} =
    \begin{cases}
        \mathcal{Z}^\top \mathcal{Z}, & \text{if } N \ge D, \\
        \mathcal{Z}\mathcal{Z}^\top,  & \text{if } N < D,
    \end{cases}
\end{equation}
where $\mathcal{G} \in \mathbb{R}^{D \times D}$ or $\mathbb{R}^{N \times N}$ accordingly.
The eigenvalue spectrum of $\mathcal{G}$ reflects the effective dimensionality of the representation space. As illustrated in Figure~\ref{fig:3dplot}, the spectrum typically exhibits an elbow point~\cite{elbow}, beyond which eigenvalues contribute marginally to the overall variance.

Following prior work~\cite{uncomp}, we retain the top-$k$ eigenvalues
$\{\lambda_1(\mathcal{G}), \ldots, \lambda_k(\mathcal{G})\}$
before the elbow point to suppress noise and redundancy.
Let $\mathcal{G}_k$ denote the rank-$k$ truncated Gram matrix (keeping the top-$k$ eigenpairs), and define the truncated trace as
$\mathrm{tr}_k(\mathcal{G}) := \sum_{j=1}^k \lambda_j(\mathcal{G}) = \mathrm{tr}(\mathcal{G}_k)$.
We then define the truncated matrix entropy $\mathcal{H}(\mathcal{Z})$ as:
\begin{equation}
\small
\mathcal{H}(\mathcal{Z}) = -\sum_{i=1}^k p_i \log p_i,\quad
p_i := \frac{\lambda_i(\mathcal{G})}{\mathrm{tr}_k(\mathcal{G})}
= \frac{\lambda_i(\mathcal{G})}{\mathrm{tr}(\mathcal{G}_k)}.
\label{eq:tme}
\end{equation}

\begin{figure}[!htbp]
    \centering
    \setlength{\subfigheight}{2.5cm} 
    \begin{subfigure}{0.49\columnwidth}
        \centering
        \includegraphics[height=\subfigheight]{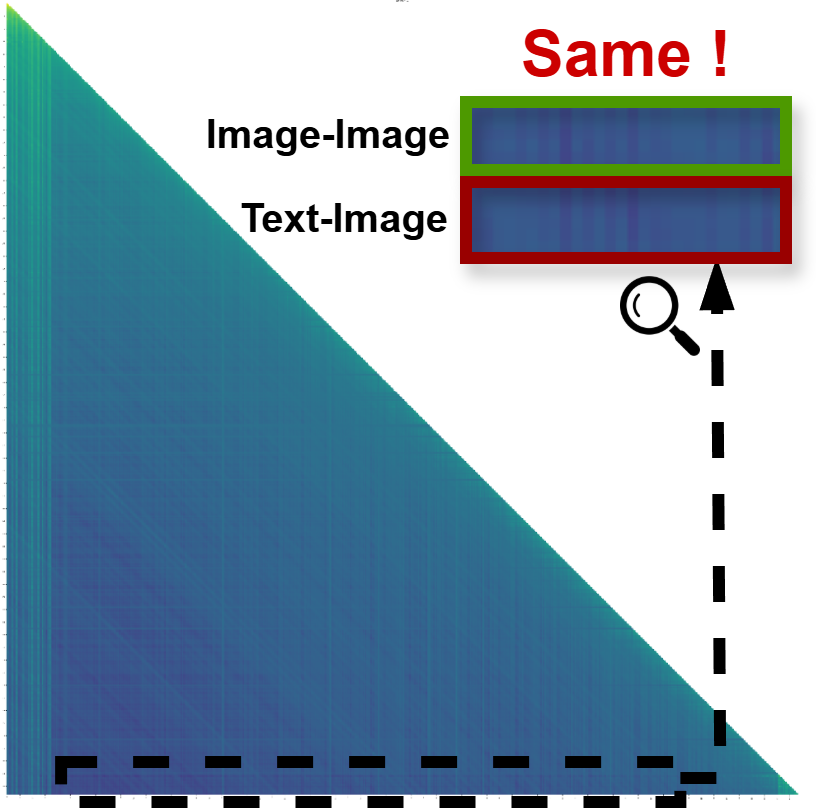}
        \caption{Initial Layer of Stage I}
        % 第一层的注意力热力图
        \label{fig:spectrum1}
    \end{subfigure} 
    \hfill
    \begin{subfigure}{0.49\columnwidth}
        \centering
        \includegraphics[height=\subfigheight]{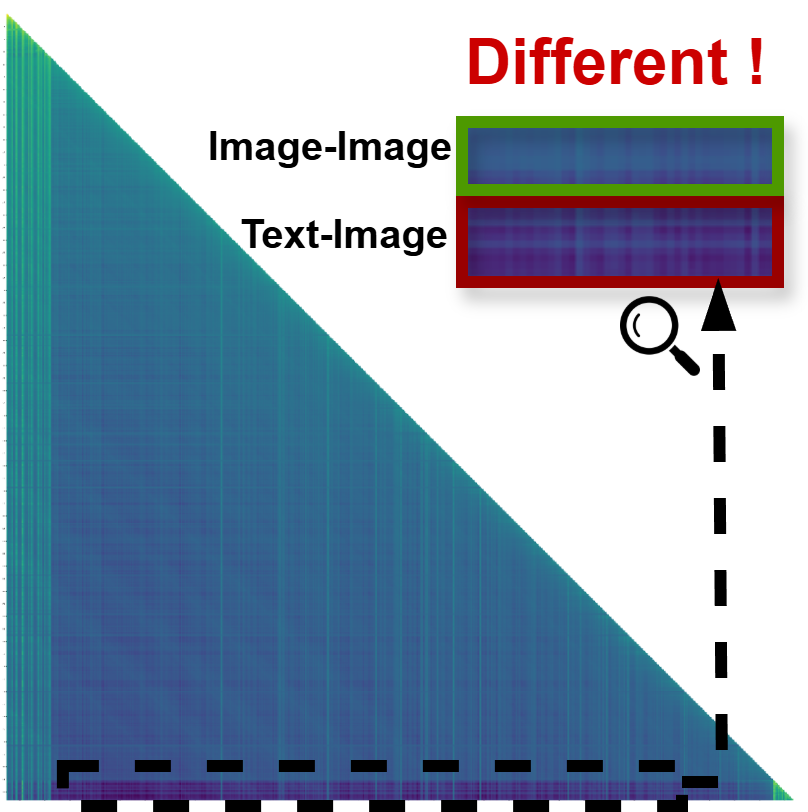}
        \caption{Final Layer of Stage I}
        % 第二层的注意力热力图
        \label{fig:spectrum2}
    \end{subfigure}
    \caption{Comparison of attention heatmaps at the initial and final layers of Stage I on LLaVA-1.5v-7B.}
    \vspace{-0.2cm}
    % At the start of Stage I, the model assigns similar attention scores to both text and visual modalities. By the end of Stage I, it differentiates the two modalities and assigns different attention scores. }
    % 从图中可以看出，在第一阶段开始时模型无法区分文本和视觉模态，因此分配相同的注意力分数；而在第一阶段结束时模型区分出两个模态，因此在不同位置分配不同的注意力分数。
    \label{fig:attention heatmap}
\end{figure}
\subsection{Unveiling the Redundancy Evolution Lifecycle} \label{sec:observation} 
\begin{figure}[t]
    \centering
    \includegraphics[width=0.9\linewidth]{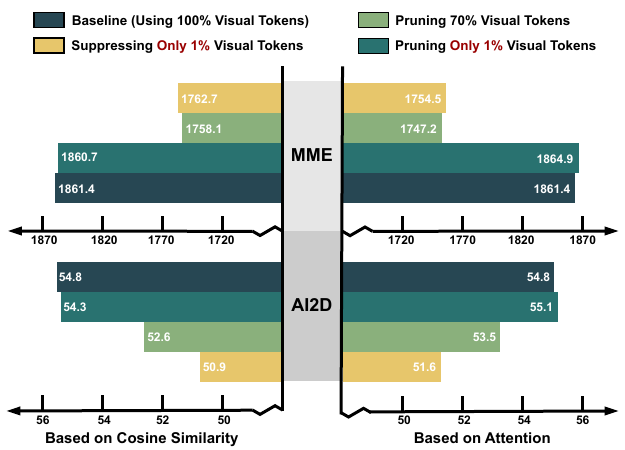}
    \caption{\textbf{Suppressing only 1\% of visual tokens leads to a performance comparable to pruning 70\% of them in Stage II.} This experiment is conducted on the MME (top) and AI2D (bottom) datasets, using cosine similarity (left) and attention scores (right) as criteria. For comparison, we also report the baseline results and the performance when pruning only 1\% of visual tokens.}
    \label{fig:locking experiment}
    \vspace{-0.4cm}
    %在第二阶段仅冻结1%的视觉token得到的性能与裁剪72%的视觉token相当。该实验分别在MME(上半部分)和AI2D(下半部分)数据集中测试，我们分别以余弦相似度(左侧)和注意力(右侧)为依据执行冻结和裁剪任务。为了便于对照，我们还提供了基线以及仅裁剪1%视觉token的评测结果。
\end{figure}
By tracking the trajectory of Truncated Matrix Entropy $\mathcal{H}(\mathcal{Z})$ across layers, we uncover a consistent three-stage redundancy lifecycle inherent to MLLM inference. This evolution delineates the distinct origins and manifestations of visual redundancy.

\paragraph{Stage I: Modality Alignment.} In early layers, we observe asymmetric entropy dynamics: visual entropy remains consistently high, whereas textual entropy rapidly compresses (see Figure \ref{fig:tme_modality}). Concurrently, attention shifts from balanced to text-dominated (see Figure \ref{fig:attention heatmap}), indicating active feature alignment. Crucially, this high-entropy visual state implies that tokens retain their raw, uncompressed structure. Consequently, the sequence is dominated by \textbf{Intrinsic Visual Redundancy (IVR)}, which refers to the spatially correlated signals \textit{\textbf{inherited}} from the ViT's dense tokenization.

\paragraph{Stage II: Global Aggregation.} As inference proceeds, the decrease in visual entropy marks the onset of global information aggregation. Our empirical analysis reveals two critical properties of this stage: 

\emph{(i) High Sensitivity to Local Perturbation.} We find that suppressing visual state updates for even a tiny fraction of tokens during this stage causes severe performance degradation (see Figure \ref{fig:locking experiment}). This confirms that the aggregation process is globally coupled, meaning any local interruption disrupts the overall integration pathway. 

\emph{(ii) Optimal Timing for One-Shot Pruning.} Given the high sensitivity, we seek the safest intervention point by evaluating the efficiency-performance of pruning. We formalize this via Marginal Utility (MU), defined as the ratio between performance drop $\Delta \mathcal{M}$ and latency gain $\Delta \mathcal{C}$:
\begin{equation}
\mathrm{MU}^{\downarrow}_{l,r}
=
\frac{-\Delta \mathcal{M}_{l,r}}{\Delta \mathcal{C}_{l,r} + \epsilon},
\end{equation}
As detailed in Table \ref{tab:mu_phase2}, minimizing this metric reveals that the onset of Stage II offers the optimal trade-off. Thus, Stage II prohibits layer-by-layer intervention but defines the precise window for one-shot IVR elimination at its boundary. 

\begin{figure*}[t]
    \centering
    \includegraphics[width=0.9\linewidth]{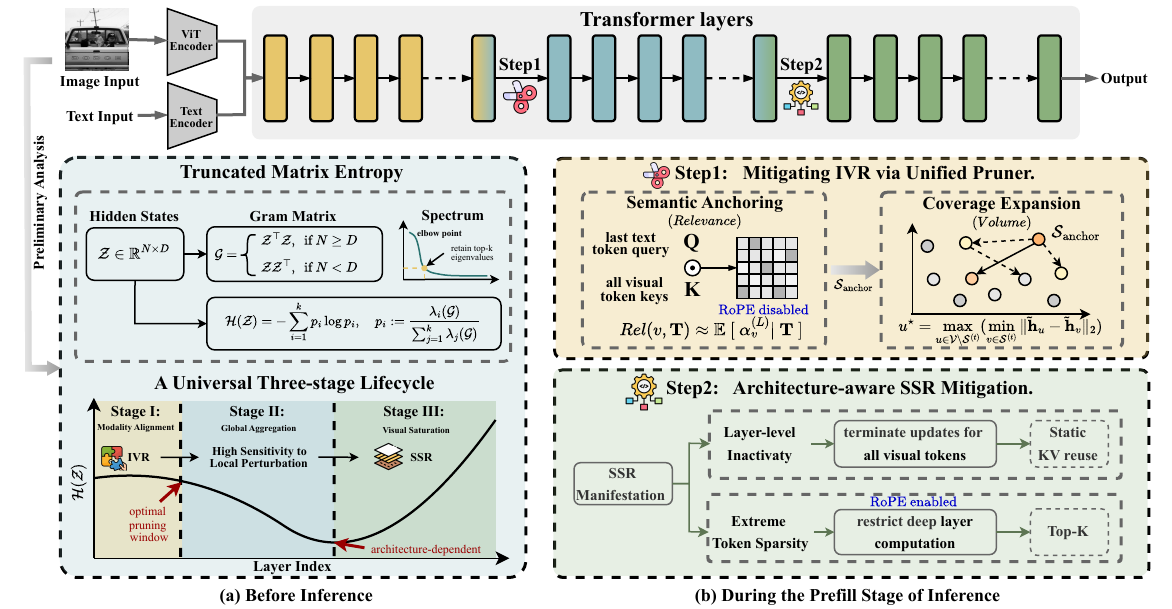}
    \caption{\textbf{Overview of HalfV.} (a) Preliminary analysis. We use a small subset of the dataset (100 samples) to identify the onset of the three internal stages in the LLM. (b) Prefill acceleration. Implementation details of our two-step acceleration strategy in the LLM prefill stage.}
    % 左侧：模型在Prefill前向传递的三个阶段中注意力矩阵形状的演进过程。值得注意的是，第三阶段被上锁的部分不参与实际的计算过程，图中可视化只是为了便于与前两个阶段进行对照。右上侧：模型在prefill前向传递的完整过程及对应的阶段。右下部份：各阶段的具体实现流程。
    %(a) 前置工作。利用少量数据集定位LLM内部三阶段的起始位置；(b)在LLM预填充阶段的两步加速策略实现细节。
    \label{fig:main}
    \vspace{-0.4cm}
\end{figure*}
%仍然需要再修改把逻辑讲通
\begin{table}[t]
\centering
% \small
% \setlength{\tabcolsep}{6pt}
\resizebox{\columnwidth}{!}{
\begin{tabular}{cccc}
\toprule
\textbf{Pruning Layer} & \textbf{Stage} & $\mathbf{MU}$ $\downarrow$ & \textbf{Latency Gain} $\uparrow$ \\
\midrule
Early Layer & I & 0.87 & \textbf{46.8\%} \\
Stage II Start & II & \textbf{0.21} & 41.5\% \\
Stage II Mid & II & 0.29 & 37.8\% \\
Deep Layer & III & 0.65 & 22.4\% \\
\bottomrule
\end{tabular}
}
\caption{Marginal utility (MU) of one-shot token pruning at different layers.}
\label{tab:mu_phase2}
\vspace{-0.2cm}
\end{table}

\begin{table}[t]
    \centering
    \resizebox{\columnwidth}{!}{
    \begin{tabular}{lccc}
    \toprule
    \textbf{Setting}&\textbf{GQA}&\textbf{OCRBench}&\textbf{ChartQA}\\
    \midrule
    LLaVA-1.5v-7B & 61.9&275&18.2\\
    \midrule
    (a) Withdraw&54.7(\textcolor{rgb,255:red,0;green,102;blue,51}{\textbf{-11.6\%}})&52(\textcolor{rgb,255:red,0;green,102;blue,51}{\textbf{-81.1\%}})&13.6(\textcolor{rgb,255:red,0;green,102;blue,51}{\textbf{-25.3\%}})\\
    (b) Suppress&60.9(\textcolor{rgb,255:red,0;green,102;blue,51}{\textbf{-1.6\%}})&311(\textcolor{red}{\textbf{+13.1\%}})&17.0(\textcolor{rgb,255:red,0;green,102;blue,51}{\textbf{-6.5\%}})\\
    \toprule
    % LLaVA-1.5v-13B & 63.2&336&18.2\\
    % \midrule
    % (a) Withdraw&55.6(\textcolor{rgb,255:red,0;green,102;blue,51}{\textbf{-12.0\%}})&52(\textcolor{rgb,255:red,0;green,102;blue,51}{\textbf{-84.5\%}})&13.5(\textcolor{rgb,255:red,0;green,102;blue,51}{\textbf{-25.8\%}})\\
    % (b)Lock&60.9(\textcolor{rgb,255:red,0;green,102;blue,51}{\textbf{-3.6\%}})&319(\textcolor{rgb,255:red,0;green,102;blue,51}{\textbf{-5.1\%}})&16.8(\textcolor{rgb,255:red,0;green,102;blue,51}{\textbf{-7.7\%}})\\
    % \toprule
    Qwen25-VL-7B & 60.7&848&84.0\\
    \midrule
    (c) Withdraw&59.7(\textcolor{rgb,255:red,0;green,102;blue,51}{\textbf{-1.6\%}})&698(\textcolor{rgb,255:red,0;green,102;blue,51}{\textbf{-17.6\%}})&75.4(\textcolor{rgb,255:red,0;green,102;blue,51}{\textbf{-10.2\%}})\\
    (d) Suppress&41.2(\textcolor{rgb,255:red,0;green,102;blue,51}{\textbf{-32.1\%}})&117(\textcolor{rgb,255:red,0;green,102;blue,51}{\textbf{-86.2\%}})&67.2(\textcolor{rgb,255:red,0;green,102;blue,51}{\textbf{-20.0\%}})\\
    (e) 5\% Tokens&60.6(\textcolor{rgb,255:red,0;green,102;blue,51}{\textbf{-0.1\%}})&827(\textcolor{rgb,255:red,0;green,102;blue,51}{\textbf{-2.4\%}})&83.6(\textcolor{rgb,255:red,0;green,102;blue,51}{\textbf{-0.4\%}})\\
    \bottomrule 
    \end{tabular}
    }
    \caption{\textbf{Comparative experiments on the LLaVA-1.5v-7B and Qwen25-VL-7B models.} We evaluate their performance on the GQA, OCRBench, and ChartQA datasets.}
    \label{tab:withdraw vs lock}
    \vspace{-0.4cm}
\end{table}
\paragraph{Stage III: Visual Saturation.} In deep layers, the visual context becomes saturated, inducing \textbf{Secondary Saturation Redundancy (SSR)}. Unlike the universal IVR, the physical manifestation of SSR is highly \textbf{architecture-dependent}, branching into two distinct patterns:

\emph{(i) Layer-level Inactivity (Vicuna/Mistral).} For architectures like LLaVA, we observe consistently low Kullback-Leibler divergence \citep{kl} ($\mathcal{D}_{KL} \approx 0$) in deep layers (Figure \ref{fig:kl}), implying minimal information gain. Experimentally, suppressing all visual updates maintains baseline performance (e.g., +13.1\% on OCRBench). Here, SSR manifests as entire layers becoming redundant, allowing for static KV reuse. 

\emph{(ii) Extreme Token Sparsity (Qwen).} In contrast, Qwen maintains high divergence, and total suppression causes catastrophic failure (e.g., -86.2\% on OCRBench). However, we find that the effective information flow collapses onto a minimal subset of dominant tokens $\mathcal{S}_{top}$. Restricting computation to just this top 5\% subset ($\mathcal{S}_{top} \subset \mathcal{V}$) restores near-lossless performance. This indicates that while layers remain active, the redundancy manifests as extreme sparsity within the token sequence. 

\begin{table*}[t]
\centering
\resizebox{\textwidth}{!}{
\begin{tabular}{c|c|cccccccc|c}
\toprule
\textbf{Method} & \textbf{TFLOPS}&\textbf{GQA} & \textbf{MME} & \textbf{POPE} & \textbf{SQA} & \textbf{VQA}$^{\textit{text}}$ & \textbf{VizWiz} & \textbf{MMB}$^{\textit{en}}$ & \textbf{AI2D} & \textbf{Average} \\
\midrule
\rowcolor{lightgraybg}\multicolumn{11}{c}{\textit{LLaVA-1.5v-7B \cite{llava}, Backbone: Vicuna}} \\

\textcolor{textgray}{Vanilla} &\textcolor{textgray}{8.31}& \textcolor{textgray}{62.0} & \textcolor{textgray}{1859} & \textcolor{textgray}{85.9} & \textcolor{textgray}{70.4} & \textcolor{textgray}{58.2} & \textcolor{textgray}{54.4} & \textcolor{textgray}{64.8} & \textcolor{textgray}{54.8} & \textcolor{textgray}{100.0\%} \\
\midrule
FastV \cite{fastv}
&3.48& 57.6 & 1730 & 81.0 & 68.9 & 52.5 & 51.3 & 61.6 & 49.7 & 94.6\% \\
VTW \cite{vtw}
&4.48& 58.7 & 1792 & 83.4 & \underline{69.6} & 49.7 & 51.8 & 63.0 & \underline{55.4} & 96.7\% \\
FitPrune \cite{fitprune}
&3.16& 58.5 & 1776 & 77.9 & 68.0 & 57.4 & 51.7 & 62.7 & 52.4 & 96.6\% \\
HiRED \cite{HiRED}
&-&58.7&1737&82.8&68.4& 47.4 &50.1&62.8&51.7&93.9\% \\
SparseVLM \cite{sparsevlm}
&3.48& 59.5 & 1787 & 85.3 & 68.6 & 56.1 & 51.4 & 60.0 & 53.5 & 96.7\% \\
PDrop \cite{pdrop}
&3.38& 57.1 & 1664 & 82.3 & 68.3 & 56.1 & 51.0 & 61.1 & 50.6 & 94.0\% \\
VisionZip \cite{visionzip}
&3.16& 59.3 & 1782 & 85.3 & 68.9 & 57.3 & 52.0 & 63.4 & 53.4 & 97.2\% \\
DivPrune \cite{divprune}
&3.16& 58.8 & 1792 & 85.1 & 68.4 & 56.8 & 51.8 & 62.1 & 52.2 & 97.1\% \\
BTP\cite{btp} 
&2.62& 59.0 & 1821 & \underline{85.6} & 69.1 & - & \underline{52.8} & 62.7 & 53.1 & 97.8\% 
\\
DART\cite{dart} 
&3.48& \underline{60.0} & \underline{1840} & 82.8 & 69.8 & 57.4 & 51.2  & 63.6 & 53.9 & 98.2\% \\
HoloV \cite{holov}&
3.16&59.0&1820&85.6&69.8& \underline{57.4} &50.9&63.9&54.1&97.8\%\\
ShortV$^{\dagger}$\cite{shortv} 
&3.77& 59.8 & 1839 & 84.0 & 68.7 & 57.1 & \underline{52.8}  & \textbf{64.8} & 53.7 & 98.4\% \\
\rowcolor{lightredbg}\textbf{HalfV (Ours)} 
&3.12& \textbf{60.5} & \textcolor{textred}{\textbf{1862}} &  \textcolor{textred}{\textbf{86.0}} & \textbf{70.2} & \textbf{57.6} & \textbf{52.9} & \underline{63.7} & \textcolor{textred}{\textbf{55.6}} & \textbf{99.2\%} \\
\midrule
\rowcolor{lightgraybg}\multicolumn{11}{c}{\textit{LLaVA-1.5v-13B \cite{llava}, Backbone: Vicuna}} \\
\textcolor{textgray}{Vanilla} &\textcolor{textgray}{16.21}& \textcolor{textgray}{63.2} & \textcolor{textgray}{1818} & \textcolor{textgray}{85.9} & \textcolor{textgray}{72.9} & \textcolor{textgray}{60.1} & \textcolor{textgray}{56.7} & \textcolor{textgray}{68.7} & \textcolor{textgray}{59.5} & \textcolor{textgray}{100.0\%} \\
\midrule
FastV \cite{fastv}
&6.68& 60.0 & 1752 & 83.6 & 72.9 & 54.7 & 53.2  & 67.0 & 56.2 & 94.6\% \\
% VTW \cite{vtw}
% &6.18& 55.6 & 1797 & 79.0 & 72.2 & 53.8 & 54.1  & 67.7 & 57.4 & \% \\
PDrop \cite{pdrop}
&5.90&60.5&1773&85.1&\underline{73.7}& 57.2 &54.2&67.3&56.9&96.0\%\\
DivPrune \cite{divprune}
&6.18& 58.8 & 1741 & 85.3 & 72.6 & \textbf{58.4} & 53.8  & 65.8 & 57.7 & 96.2\% \\
BTP \cite{btp} 
&5.72& \textbf{62.2} & 1819 & \underline{86.1} & 72.7 & \underline{58.3} & \underline{54.5}  & 68.0 & 58.3 & 98.2\% \\
ShortV$^{\dagger}$ \cite{shortv} 
&7.26& \underline{62.0} & \underline{1802} & 85.7 & 73.5 & 57.9 & 54.1 & \textbf{68.6} & \underline{58.9} & 98.6\% \\
\rowcolor{lightredbg}\textbf{HalfV (Ours)} 
&4.81& 61.6 & \textcolor{textred}{\textbf{1831}} & \textcolor{textred}{\textbf{87.2}} & \textcolor{textred}{\textbf{75.0}} & 58.0 & \textbf{55.1}  & \underline{68.5} & \textbf{59.3} & \textbf{99.4\%} \\
\midrule
\rowcolor{lightgraybg}\multicolumn{11}{c}{\textit{LLaVA-NeXT-7B \cite{llavanext}, Backbone: Mistral}} \\
\textcolor{textgray}{Vanilla} &\textcolor{textgray}{40.42}& \textcolor{textgray}{64.8} & \textcolor{textgray}{1824} & \textcolor{textgray}{86.8} & \textcolor{textgray}{78.7} & \textcolor{textgray}{65.8} & \textcolor{textgray}{63.8}  & \textcolor{textgray}{68.1} & \textcolor{textgray}{67.3} & \textcolor{textgray}{100.0\%} \\
\midrule
FastV \cite{fastv}
&15.03& 60.3 & 1782 & 85.5 & 70.1 & 59.7 & 58.4  & 64.3 & 66.1 & 94.1\% \\
% PDrop \cite{pdrop}
% &14.62& 60.9 & 1784 & 85.9 & 72.4 & 62.1 & 59.7  & 65.1 & 67.0 & \% \\
DivPrune \cite{divprune}
&13.34& 61.4 & 1703 & 86.2 & 71.9 & 63.7 & 60.0 & 65.4 & 66.5 & 95.7\% \\
BTP \cite{btp} 
&11.92& 60.6 & 1802 & \underline{86.7} & 74.4 & 63.2 & 61.8 & 66.3 & 65.7 & 97.1\% \\
DART \cite{dart}&15.03&62.4&1789&86.0&75.3& \textbf{64.5} &61.7&\underline{67.5}&65.9&97.7\%\\
HoloV \cite{dart}&13.34&62.5&1796&85.9&76.2& 64.0 &\underline{62.2}&67.1&66.5&97.8\%\\
ShortV$^{\dagger}$ \cite{shortv} 
&16.92& \textbf{63.4} & \underline{1809} & 86.5 & \underline{77.1} & 63.9 & 61.4 & 67.2 & \underline{66.7} & 97.9\% \\
\rowcolor{lightredbg}\textbf{HalfV (Ours)} 
&11.22& \underline{62.5} & \textbf{1813} & \textcolor{textred}{\textbf{87.3}} & \textcolor{textred}{\textbf{78.8}} & \underline{64.2} & \textbf{62.4} & \textbf{68.0} & \textbf{67.0} & \textbf{98.9\%} \\
\midrule
\rowcolor{lightgraybg}\multicolumn{11}{c}{\textit{Qwen25-VL-7B \cite{qwen25vl}, Backbone: Qwen2.5}} \\
\textcolor{textgray}{Vanilla} &\textcolor{textgray}{36.63}& \textcolor{textgray}{60.7} & \textcolor{textgray}{2307} & \textcolor{textgray}{86.5} & \textcolor{textgray}{88.7} & \textcolor{textgray}{83.2} &\textcolor{textgray}{70.6} & \textcolor{textgray}{82.5} & \textcolor{textgray}{81.4} & \textcolor{textgray}{100.0\%} \\
\midrule
FastV \cite{fastv}
&10.33& 52.7 & 2036 & 80.7 & 78.0 & 72.5 & 61.2  & 74.9 & 67.1 & 87.9\% \\
DivPrune \cite{divprune} 
&8.30& 50.1 & 2002 & 83.9 & 73.0 & 70.9 & 62.8 & 76.9 & 75.1 & 88.1\% \\
BTP \cite{btp} 
&11.16& 57.2 & \underline{2198} & \textbf{84.7} & 74.1 & 72.1 & 62.8 & 75.2 & 74.3 & 91.1\% \\
DART \cite{dart} 
&10.33& 56.5 & 2137 & 82.9 & 82.1 & 75.3 & 67.4  & \underline{79.3} & 77.8 & 93.9\% \\
HoloV \cite{holov}&8.30&54.3&2043&82.3&79.8& 70.3 &-&76.5&75.6&90.5\%\\
ShortV$^{\dagger}$ \cite{shortv} 
&13.57& 47.5 & 1516 & 68.7 & 72.8 & 64.1 & 59.8  & 74.1 & 67.4 & 80.0\% \\
\rowcolor{lightredbg}\textbf{HalfV (Ours} 
&8.91& \textbf{58.4} & \textbf{2242} & \underline{84.5} & \textbf{87.0} & \textbf{76.7} & \textbf{68.5} & \textbf{81.1} & \textbf{79.7} & \textbf{96.8\%} \\
\bottomrule
\end{tabular}
}
\caption{\textbf{Comparison with state-of-the-art inference acceleration methods on LLaVA-1.5-7B/13B, LLaVA-NeXT-7B and Qwen25-VL-7B models.} The best result in each group is highlighted in \textbf{bold}, and the second-best result is \underline{underlined}. Results that are higher than the original model are marked in \textbf{\textcolor{textred}{red}}. ${\dagger}$ method focuses on layer-level redundancy, while others focus on token-level redundancy.}
\label{tab:main_results}
\vspace{-0.4cm}
\end{table*}

\subsection{HalfV}
\label{sec:halfv}
% 早期数据冗余：
% 在上一小节中，我们发现视觉token在第二阶段进行模态内信息融合，为了保证该阶段全部视觉信息的充分融合，同时加速推理，我们选择在该阶段开始时进行无位置偏差的视觉token裁剪。从图1a中可以看出，早期层的注意力裁剪受到位置编码的影响，会对更靠后的视觉分配更高的注意力分数。在模型更深层位置编码造成的影响会下降。我们发现移除位置编码后得到的注意力分数可以更好地保留视觉的全局信息和重要信息。因此，在该阶段我们选择使用无偏差的token裁剪从数据层面减少冗余。
%基于上述发现，我们提出了一种多阶段推理加速方法——HalfV。该方法首先在模型的第二阶段起始层进行视觉token裁剪，然后在第三阶段冻结第二阶段保留的视觉token，仅将其作为文本token查询的键和值进行前向传播。

Based on the observations, we propose HalfV, a two-step inference acceleration method. The overview of HalfV is illustrated in Figure \ref{fig:main}.

\paragraph{Step1: Mitigating IVR via Unified Pruner.}
We perform token pruning at the boundary of Stage~I and Stage~II. Our goal is  to select a subset $\mathcal{S}$ that preserves the cross-modal semantic alignment established in Stage~I while mitigating IVR. Crucially, pruning should avoid representational collapse so that the retained tokens provide a stable basis for Stage~II global aggregation. Given visual tokens $\mathcal{V}$ and a budget $K$, we denote the optimal subset by $\mathcal{S}^{\star}$ and formulate:
% 我们的目标是选取一个子集S，使其在保持 Stage~I 语义对齐与缓解内在视觉冗余（IVR）之间取得平衡；特别是，我们要求在去冗余的过程中避免空间塌缩，从而为 Stage~II 的全局聚合奠定稳健的表示基础。在给定预算K的条件下，我们将满足上述要求的最优子集记为S^{star}，并将其形式化为如下双目标最大化问题：
\begin{equation}
\small
\mathcal{S}^\star
= \argmax_{\mathcal{S} \subset \mathcal{V},\, |\mathcal{S}|=K}
\left(
\sum_{v \in \mathcal{S}} \mathrm{Rel}(v, \mathbf{T})
+ \lambda \cdot \mathrm{Vol}(\mathcal{S})
\right).
\label{eq:subset_obj}
\end{equation}
The first term $\sum_{v \in \mathcal{S}} \mathrm{Rel}(v, \mathbf{T})$ aggregates token-wise relevance to the textual condition $\mathbf{T}$, while the second term $\mathrm{Vol}(\mathcal{S})$ promotes geometric spread (coverage) in feature space to prevent degeneration; $\lambda>0$ balances the two. Since Eq.~\eqref{eq:subset_obj} is NP-hard, we propose \textbf{AnchorCover}, a greedy solver that optimizes the two terms sequentially.

% The first term $\sum_{v \in \mathcal{S}} \mathrm{Rel}(v, \mathbf{T})$ aggregates the semantic relevance of each visual token $v$ to the textual condition $\mathbf{T}$. The second term $\mathrm{Vol}(\mathcal{S})$ promotes geometric spread (coverage) in feature space to prevent degeneration. The scalar $\lambda>0$ balances relevance and geometric coverage.

% \paragraph{The SeedCover Solver.}
% 由于直接优化式 \eqref{eq:subset_obj} 属于 NP-hard 的组合优化问题，我们提出 \textbf{SeedCover}，以两个串行的贪心步骤对双目标进行近似求解，从而高效逼近全局最优。

% \textit{1) 相关性最大化（Anchoring）。}
% 第一阶段通过最大化 $\sum_{v\in\mathcal{S}}\mathrm{Rel}(v,\mathbf{T})$ 锁定语义锚点。我们采用跨模态注意力权重 $\alpha^{(L)}$ 作为可计算的 token 效用代理，即 $\mathrm{Rel}(v,\mathbf{T}) \approx \mathbb{E}[\alpha^{(L)}v \mid \mathbf{T}]$。为兼顾硬件友好性与全局条件信息，我们仅使用\textbf{末位文本 token} 的 Query 与所有视觉 token 的 Key 计算交叉注意力：这是因为末位 token 已汇聚完整指令语义，能够提供充分的条件表征；同时该计算可简化为一次向量积，从而天然适配 FlashAttention 等加速算子。消融实验表明，该模块额外开销仅占 TTFT 的 0.7%。此外，为削弱位置偏差对相关性估计的干扰，我们在计算该代理分数时禁用位置编码（如 RoPE）。最终，选取注意力得分最高的 Top-$K_s$ 个视觉 token 构成\textbf{锚点集合} $\mathcal{S}{\text{anchor}}$，以极低开销优先保留任务关键线索。
% Directly optimizing Eq.~\eqref{eq:subset_obj} is NP-hard. We propose \textbf{AnchorCover}, a greedy algorithm that approximates the global optimum by addressing the two terms sequentially.

(1) Relevance Maximization (Semantic Anchoring).
We use cross-modal attention $\alpha^{(L)}$ as a utility proxy where $L$ is the pruning layer, i.e., $\mathrm{Rel}(v,\mathbf{T}) \approx \mathbb{E}[\alpha^{(L)}_v|\mathbf{T}]$. For hardware efficiency and conditioned relevance, we compute cross-attention using only the query of the last textual token against the keys of all visual tokens. The last token summarizes the instruction, and the computation reduces to a single vector product compatible with FlashAttention \cite{flashattention}. As shown in Tab.~\ref{tab:albation_module}, this module adds negligible overhead (0.7\% TTFT). To reduce positional bias, we disable position embeddings \cite{rope} when computing scores. We then form $\mathcal{S}_{\text{anchor}}$ by selecting the Top-$K_{\mathcal{S}}$ tokens.

(2) Volume Maximization (Coverage Expansion).
With the relevant anchors fixed, we further aims to maximize the incremental geometric coverage $\Delta\text{Vol}(\mathcal{S})$.
It is a known geometric property that the volume of a simplex is maximized when its vertices are mutually distant.
Therefore, we employ \emph{Farthest Point Sampling (FPS)} \cite{fps} as a greedy surrogate. Let $\tilde{\mathbf{h}}_v$ denote the $\ell_2$-normalized hidden state of token $v$.
At each step, we select the token $u^\star$ that is most distant from the current set $\mathcal{S}^{(t)}$:
\begin{equation}
\small
u^\star = \argmax_{u \in \mathcal{V} \setminus \mathcal{S}^{(t)}} \left( \min_{v \in \mathcal{S}^{(t)}} \| \tilde{\mathbf{h}}_u - \tilde{\mathbf{h}}_v \|_2 \right).
\end{equation}
We iterate $u^\star$ into $\mathcal{S}^{(t+1)}$ until $|\mathcal{S}^{(t)}| = K$, obtaining a coverage set $\mathcal{S}_{\text{cover}}$. The final selection
\begin{equation}
\small
\mathcal{S}^\star = \mathcal{S}_{\text{anchor}} \cup \mathcal{S}_{\text{cover}}
\end{equation}
By prioritizing tokens that are nearly orthogonal to the existing anchors, this step effectively mitigates the spatial redundancy while ensuring the final representation forms a robust basis for Stage~II.

\paragraph{Step2: Architecture-aware SSR Mitigation.}We adopt differentiated strategies based on SSR manifestations.

(1) Handling Layer-level Inactivity. For architectures like LLaVA, we terminate both attention and feed-forward updates for all visual tokens ($\Delta\mathbf{H}_V = \mathbf{0}$) to eliminate redundant computation, while maintaining normal updates for textual tokens:
\vspace{-0.2cm}
\begin{equation}
\small
\begin{split}
\Delta\mathbf{H}_V^{(l)} &= \mathbf{0}, \\
\Delta\mathbf{H}_T^{(l)} &= \mathcal{F}(\mathbf{H}_T^{(l)}, \operatorname{Concat}(\mathbf{H}_V^{(l)}, \mathbf{H}_T^{(l)})).
\end{split}
\label{eq:freeze}
\vspace{-0.2cm}
\end{equation}
More details are provided in Appendix \ref{appendix:layer_inactivaty}.

(2) Handling Extreme Token Sparsity. For architectures like Qwen, we sparsify deep-layer computation by updating only the Top-$K_\text{SSR}$ visual tokens. Concretely, we rank visual tokens by the cross-modal attention scores induced by the last textual token with positional embeddings (e.g., RoPE) enabled, and select $\mathcal{S}_{top}$ accordingly. Computation in subsequent layers is then restricted to tokens in $\mathcal{S}_{top}$, while the remaining visual tokens are pruned. This design differs from our IVR stage, where positional encoding is disabled for scoring. We discuss the rationale and ablations in Appendix~\ref{appendix:rope}.

\section{Experiments}
\subsection{Experimental Settings}
% To comprehensively assess the generalization capability of our approach, we integrate it into multiple state-of-the-art multimodal large models and conduct extensive evaluations on a diverse set of benchmark tasks. In particular, we validate our method on four representative models with three different backbones: LLaVA-v1.5-7B/13B (Vicuna), LLaVA-v1.6-7B (Mistral), and Qwen2.5-VL-7B (Qwen2.5) \citep{llava,llavanext,qwen25vl}. Implementation details are provided in Appendix \ref{sec:Experiment Settings}.
We conduct extensive evaluations to verify the robustness of HalfV across varying model architectures. Our benchmarks cover four mainstream MLLMs built upon different linguistic foundations: the Vicuna-based LLaVA-1.5v-7B/13B \cite{llava}, the Mistral-based LLaVA-1.6v-7B \cite{llavanext}, and the Qwen-based Qwen2.5-VL-7B~\citep{qwen25vl}. For comprehensive implementation details, please refer to Appendix~\ref{sec:Experiment Settings}.
% In particular, we assess our approach on four prominent models: LLaVA-v1.5-7B, LLaVA-v1.5-13B \cite{llava}, LLaVA-NeXT-7B \cite{llavanext}, and Qwen2.5-VL 7B-Instruct \cite{qwen25vl}.
\subsection{Main Results}
\begin{table}[t]
\centering
\setlength{\tabcolsep}{4pt}       % 调整列间距
\resizebox{\columnwidth}{!}{
\begin{tabular}{l|ccccccc|c}
\toprule
\textbf{Method} & \textbf{GQA} & \textbf{MMB} & \textbf{MME} & \textbf{POPE} & \textbf{SQA} & \textbf{VQA$^{\textit{Text}}$}&\textbf{AI2D} & \textbf{Avg.} \\

% === 第一部分: Upper Bound ===
\midrule
\rowcolor{lightgraybg} 
Qwen2-VL-7B & \multicolumn{8}{c}{\textit{Upper Bound, All Tokens} (\textbf{100\%})} \\
\textcolor{textgray}{Vanilla} & \textcolor{textgray}{60.7} & \textcolor{textgray}{82.5} & \textcolor{textgray}{2307} & \textcolor{textgray}{86.5} & \textcolor{textgray}{88.7} & \textcolor{textgray}{83.2} & \textcolor{textgray}{81.4} & \textcolor{textgray}{100\%} \\

% === 第二部分: 66.7% Reduction ===
\hline
\rowcolor{lightgraybg} 
Qwen2-VL-7B & \multicolumn{8}{c}{\textit{Flops Ratio Reduction} (\textbf{$\downarrow$ 88.9\%})} \\
+ FastV \textit{(ECCV24)} & 50.1 & 69.2 & 1940 & 78.6 & 77.4 & 60.3 &68.7 & 83.6\% \\
+ DART \textit{(EMNLP25)}& 54.3 & 74.8 & 2086 & 76.9 & 81.7 & 62.5 &72.1 & 87.9\% \\
+ HoloV \textit{(NIPS25)}& 52.8 & 72.4 & 2006 & 80.7 & 79.5 & 61.8 &72.6& 86.8\% \\
\rowcolor{lightredbg}\textbf{+ HalfV (Ours)} & \textbf{56.8} & \textbf{79.2} & \textbf{2124} &\textbf{82.6} & \textbf{86.5} & \textbf{68.8} & \textbf{74.6} & \textbf{92.7\%} \\

\midrule
\rowcolor{lightgraybg} 
Qwen3-VL-4B-FP8 & \multicolumn{8}{c}{\textit{Upper Bound, All Tokens} (\textbf{100\%})} \\
% mmb 88.5  sqa:92
\textcolor{textgray}{Vanilla} & \textcolor{textgray}{59.0} & \textcolor{textgray}{83.6} & \textcolor{textgray}{-} & \textcolor{textgray}{85.7} & \textcolor{textgray}{90.8} & \textcolor{textgray}{81.7} & \textcolor{textgray}{83.9} & \textcolor{textgray}{100\%} \\
% === 第三部分: 77.8% Reduction ===
\hline
\rowcolor{lightgraybg} 
Qwen3-VL-4B-FP8 & \multicolumn{8}{c}{\textit{Flops Ratio Reduction} (\textbf{$\downarrow$ 77.8\%})} \\
+ FastV \textit{(ECCV24)}& 53.7 & 77.5 & - & 78.9 & 84.1 & 70.3 &77.5& 91.1\% \\
+ DART \textit{(EMNLP25)}& 55.8 & 79.2 & - & 82.5 & 85.9 & 72.2 &80.3& 94.0\% \\
\rowcolor{lightredbg}\textbf{+ HalfV (Ours)} & \textbf{57.3} & \textbf{80.9} & - &\textbf{83.1} & \textbf{87.5} & \textbf{76.4} & \textbf{81.4}  & \textbf{96.3\%} \\

% === 第四部分: 88.9% Reduction ===
\hline
\rowcolor{lightgraybg} 
Qwen3-VL-4B-FP8 & \multicolumn{8}{c}{\textit{Flops Ratio Reduction} (\textbf{$\downarrow$ 88.9\%})} \\
+ FastV \textit{(ECCV24)}& 49.7 & 74.9 & - & 74.6 & 82.7 & 66.7&74.2& 87.0\% \\
+ DART \textit{(EMNLP25)}& 52.4 & 76.2 & - & 78.3 & 83.9 & 70.1 &76.3 & 90.1\% \\
\rowcolor{lightredbg}\textbf{+ HalfV (Ours)} & \textbf{54.1} & \textbf{77.4} & - &\textbf{79.2} & \textbf{84.7} & \textbf{75.2} & \textbf{78.1} & \textbf{92.5\%} \\
\bottomrule
\end{tabular}
}
\caption{More comparative experiments on Qwen2-VL-7B and Qwen3-VL-4B models.}
\label{tab:qwen3}
\vspace{-0.2cm}
\end{table}
\begin{table}[t]
    \centering
    \resizebox{\columnwidth}{!}{
    \begin{tabular}{lcccccccc}
    \toprule
     \multirow{2}{*}{\textbf{Method}}&\multicolumn{2}{c}{\textbf{TGIF}}&\multicolumn{2}{c}{\textbf{MSVD}}&\multicolumn{2}{c}{\textbf{MSRVT}}& \multicolumn{2}{c}{\textbf{Avg.}}\\
     \cmidrule(lr){2-3}
    \cmidrule(lr){4-5}
    \cmidrule(lr){6-7}
    \cmidrule(lr){8-9}
    &Acc.&Score&Acc.&Score&Acc.&Score&Acc.&Score\\
    \midrule
    VideoChat-7B &34.4 &2.3 &56.3 &2.8& 45.0& 2.5& 45.1& 2.5\\
    LLaMA-Adapter-7B&-&-&54.9& 3.1& 43.8& 2.7&-&-\\
    Video-ChatGPT& 51.4 &3.0 &64.9 &3.3& 49.3& 2.8& 55.2& 3.0\\
    \midrule
    Video-LLaVA-7B& 47.0 &3.4& 70.2& 3.9& 57.3 &3.5& 58.2 &3.6\\
    + FastV (\textit{ECCV24})& 45.2 &3.1& \textbf{71.0} &3.9&55 &3.5& 57.1 &3.5\\
    + DART (\textit{EMNLP25})&46.3 &3.4& \textbf{71.0} &4.0& 56.7 &3.6& 58.0 &3.7\\
    + HoloV (\textit{NIPS25})&46.1&3.2&\textbf{71.0}&4.0&56.5&3.6&57.8&3.6\\
    \rowcolor{lightredbg}\textbf{+ HalfV (Ours)}&\textbf{46.5}&\textbf{3.5}&\underline{70.8}&\textbf{4.0}&\textbf{56.9}&\textbf{3.6}&\textbf{58.1}&\textbf{3.7}\\
    \bottomrule
    \end{tabular}
    }
    \caption{Comparative experiments on video understanding tasks.}
    \label{tab:video}
    \vspace{-0.4cm}
\end{table}
%BTP 在各类 LVLM 上优于现有 SOTA 方法。 如表 1 所示，我们在不同模型家族与不同参数规模下进行了大量实验。实证结果表明，我们的方法在大多数基准任务上都能稳定超过当前最先进的方法。在不同规模的 LLaVA 模型上，即使在 22% 的压缩率 下，我们的方法仍能保持 原始平均性能的 98%。此外，我们的方法在各个模型上都持续优于其他方法，其结果同时超过基于注意力的方法与基于多样性的方法。我们还在图 5 中可视化了不同方法对各层输出的影响：我们的方法在局部与全局层面都能保持与原始输出的一致性。附录 7.5 进一步提供了各方法所选择的图像 token 的空间分布可视化。结果显示，我们的方法在更深层能够实现更有效的 token 选择。
\noindent\textbf{HalfV Maintains High Performance across Diverse MLLM Backbones.} As shown in Table \ref{tab:main_results}, our method achieves SOTA or near-SOTA results across various model series and scales. Notably, on the Qwen series, HalfV significantly outperforms previous SOTA methods: (i) HalfV retains \textbf{96.8\%} performance at 4.1$\times$ FLOPs speedup, outperforming DART by \textbf{2.9\%}; and (ii) at 6.1$\times$ FLOPs speedup, this gap widens to \textbf{4.8\%} (Table \ref{tab:qwen3}). Experiments on Qwen3-VL-4B-FP8 \cite{qwen3} further validate our superiority across different acceleration ratios. For video understanding, integrating HalfV into Video-LLaVA \citep{videollava} (Table \ref{tab:video}) yields performance comparable to the original model, demonstrating strong reasoning retention even with high-resolution inputs.

% \noindent\textbf{HalfV Mitigates Hallucination and Surpasses the Baseline.} HalfV not only preserves accuracy on hallucination benchmarks for the LLaVA series, but also improves upon the uncompressed baseline on multiple metrics. Prior inference acceleration methods typically aim to preserve the original accuracy under high speedup ratios. In contrast, our results indicate that HalfV can provide additional gains on hallucination-related evaluations, yielding accuracy that is even higher than the original (non-accelerated) model. As shown in Table \ref{tab:main_results}, on LLaVA-1.5v-7B, HalfV achieves 86.0\% accuracy on POPE, exceeding the baseline’s 85.9\% and also delivers better MME scores. On LLaVA-1.5v-13B, HalfV further improves POPE by +1.3\% over the baseline, and obtains 1831 on MME, improving by +13 points. Similar gains are observed on LLaVA-NeXT-7B. We attribute these improvements to suppressing IVR and SSR during inference, which reduces misleading visual evidence and enhances prediction reliability.
\noindent\textbf{HalfV Mitigates Hallucination and Surpasses Baseline.} Unlike prior methods that merely aim to preserve accuracy, HalfV surpasses the uncompressed baseline on hallucination benchmarks. As shown in Table \ref{tab:main_results}, HalfV improves POPE on LLaVA-1.5v-7B (86.0\% vs. 85.9\%) and further boosts LLaVA-1.5v-13B by \textbf{+1.3\%} on POPE and \textbf{+13} points on MME (reaching 1831). Similar gains are observed on LLaVA-NeXT-7B. We attribute these improvements to the suppression of IVR and SSR, which effectively reduces misleading visual evidence.
% HalfV在不同骨架的MLLM上都可以保持高性能。如表1所示，我们在不同模型系列、骨架与不同参数规模下进行了大量实验。实验结果表明，我们的方法在大多数基准任务上都能实现最优或次优的结果。值得注意的是，我们的方法在Qwen系列模型上取得的性能显著优于其他方法。(i) 在4.1倍的压缩率下，HalfV仍能保持96.8\%的平均性能，较次优方法DART显著领先2.9\%。(ii) 在更激进的加速策略下，这一优势进一步扩大：如表2所示，当我们将压缩率提升至6.1倍时，HalfV的优势扩大至4.8\%。此外，我们还在Qwen3-VL-4B-FP8模型中测试了我们的方法。实验结果如表2所示，在不同的加速比例下，我们的方法均优于其他的方法。在视觉理解基准上，我们将HalfV集成到Video-LLaVA中，并将其与主流的集中加速方法进行对比评测。如表3所示，HalfV的性能与原始模型保持很接近。这表明在处理高分辨率视觉输入时，HalfV仍然能够保持极强的推理能力。

% HalfV在处理LLaVA系列模型的的幻觉任务时性能超过基础模型。对于推理加速领域的工作聚焦于如何实现高加速比的同时维持模型原本的性能。然而，我们发现HalfV在处理幻觉评测任务基线上，其准确率甚至可以超越基础模型。如表1所示，对于LLaVA-1.5v-7B模型，其在POPE基准中可以取得86.0\%的准确率，高于基础模型的85.9\%的准确率。其在MME基准中可以取得1862\%的分数，比基础模型分数还要高3分。对于LLaVA-1.5v-13B模型，其在POPE基准中甚至比基础模型的准确率还要高出1.3\%,同时在MME中的表现也优于基础模型。对于LLaVA-NeXT-7B模型，其在POPE基准中的准确率也高于基础模型。上述实验结果意味着，我们可以通过在模型推理过程中降低IVR和SSR两类冗余从而缓解模型的幻觉。
\begin{figure*}[t]
    \centering
    % 第1张图
    \begin{subfigure}[b]{0.245\textwidth}
        \centering
        \includegraphics[width=\textwidth]{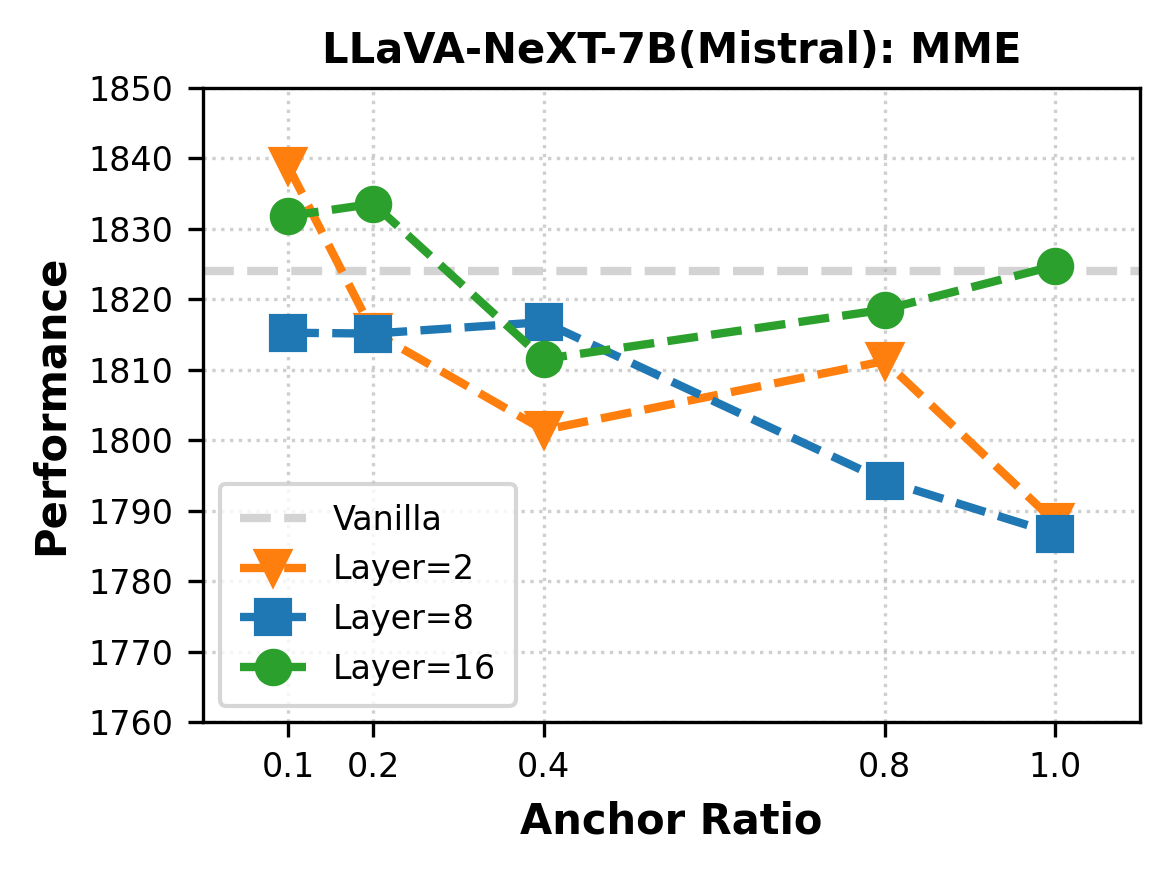}
        \label{fig:sub1}
    \end{subfigure}
    \hfill % 弹性空格，把图片撑开
    % 第2张图
    \begin{subfigure}[b]{0.245\textwidth}
        \centering
        \includegraphics[width=\textwidth]{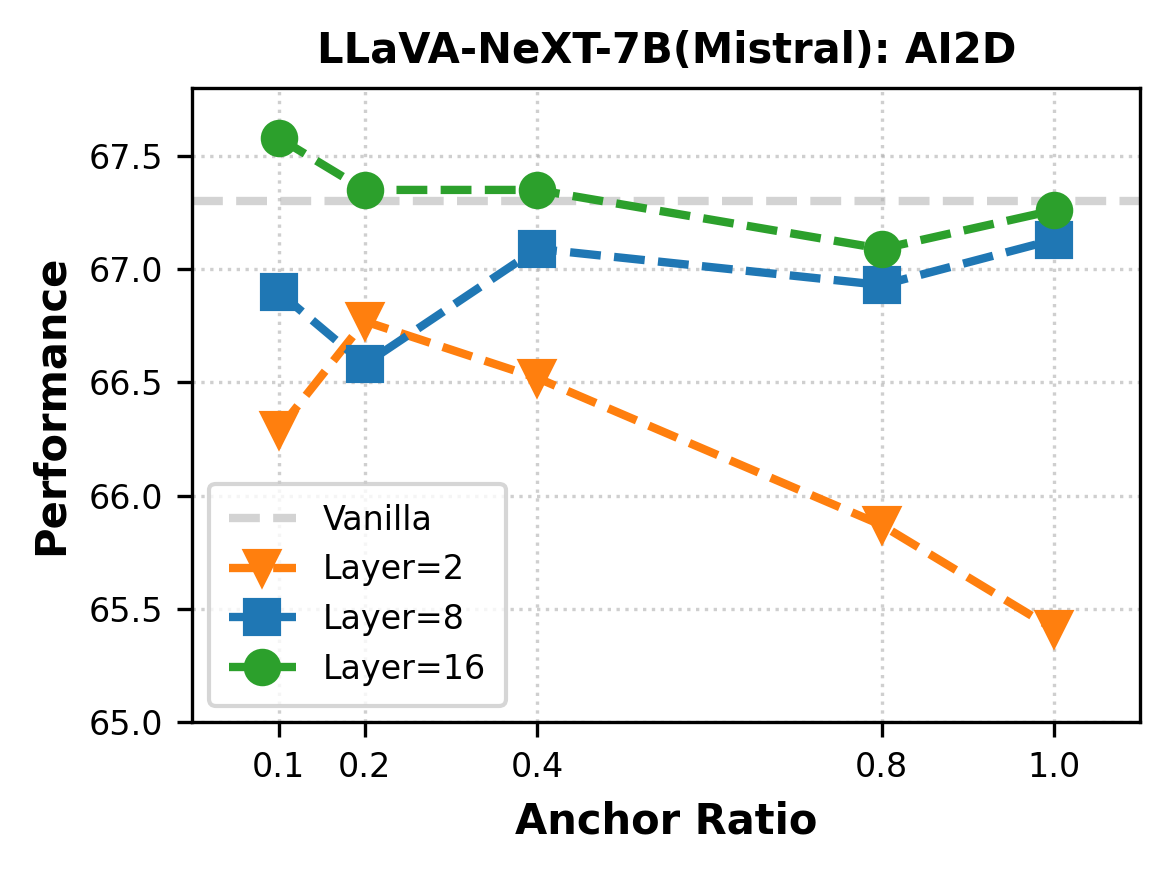}
        \label{fig:sub2}
    \end{subfigure}
    \hfill
    % 第3张图
    \begin{subfigure}[b]{0.245\textwidth}
        \centering
        \includegraphics[width=\textwidth]{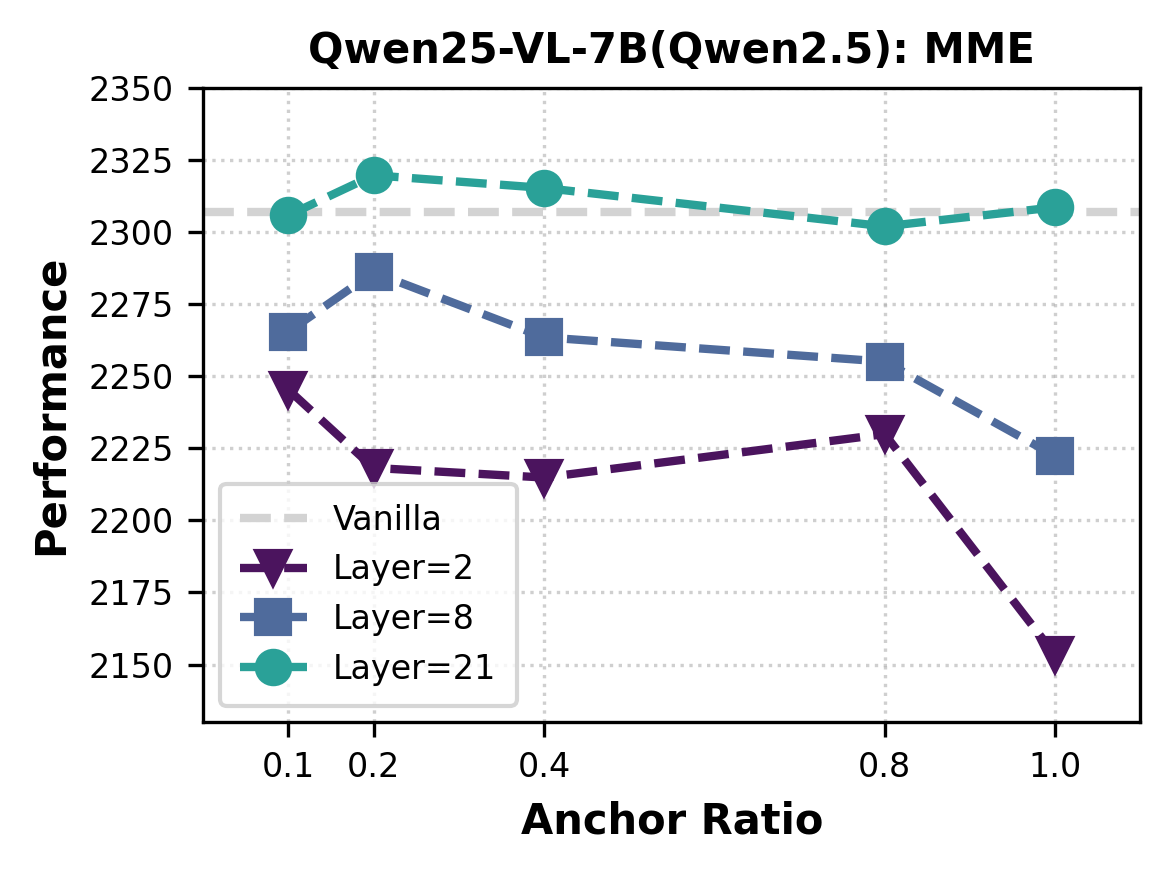}
        \label{fig:sub3}
    \end{subfigure}
    \hfill
    % 第4张图
    \begin{subfigure}[b]{0.245\textwidth}
        \centering
        \includegraphics[width=\textwidth]{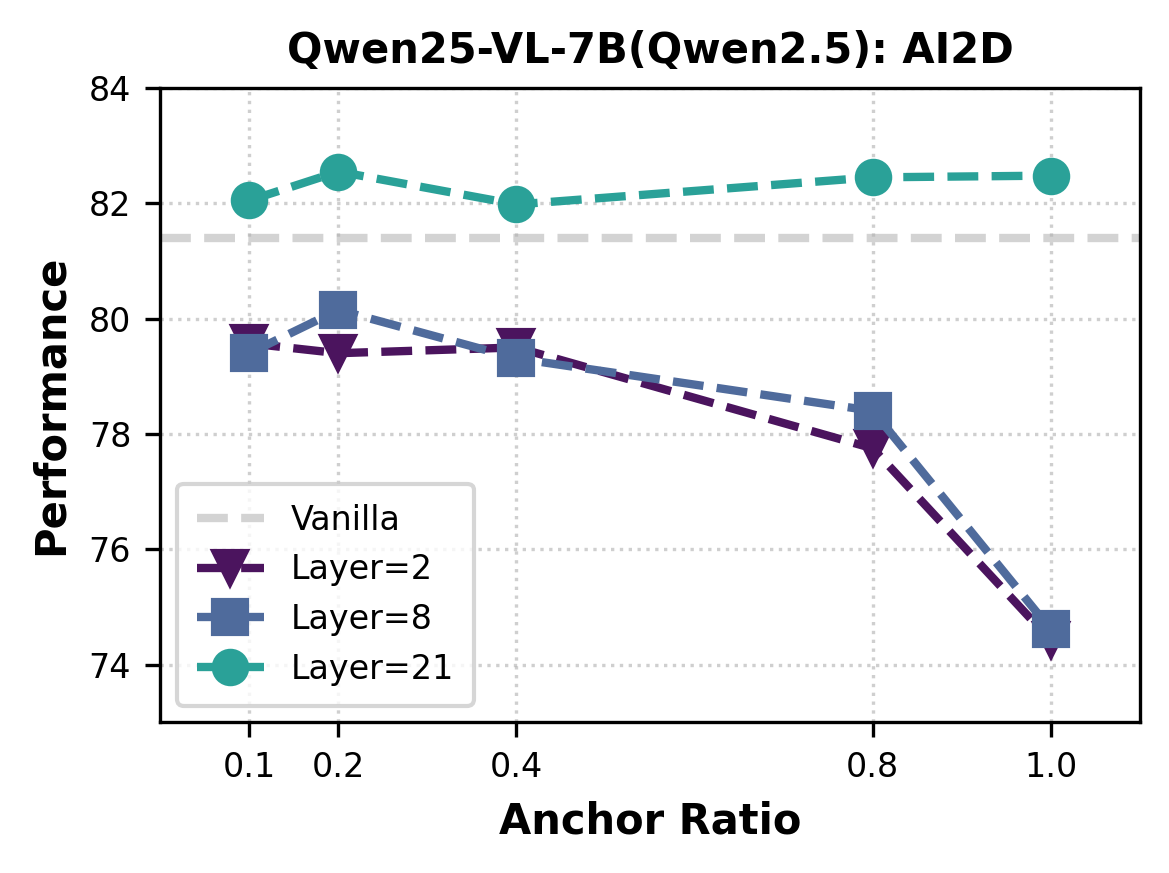}
        \label{fig:sub4}
    \end{subfigure}
    \vspace{-1.0cm}
    \caption{Performance comparison of different models under different pruning layers and anchor retention ratios.}
    \vspace{-0.4cm}
    % LLaVA-NeXT-7B和Qwen25-VL-7B模型在不同层、不同锚点率下在MME和Ai2D数据集中的性能对比图
    \label{fig:anchor_ratio}
\end{figure*}
\subsection{Efficiency Analysis}
\begin{table}[t]
    \centering
    \resizebox{\columnwidth}{!}{
    \begin{tabular}{cccccc}
    \toprule
    \multirow{2}{*}{\textbf{Method}}&\textbf{Total Time}&\textbf{TTFT}&\textbf{Prefill Time}&\textbf{Latency}&\multirow{2}{*}{\textbf{Perf.}}\\
     \cmidrule(lr){2-2}
     \cmidrule(lr){3-3}
    \cmidrule(lr){4-4}
    \cmidrule(lr){5-5}
    &\textbf{(Min:Sec)}&\textbf{(Millisecond)}&\textbf{(Millisecond)}&\textbf{(Millisecond)}&\\
    \midrule
    \rowcolor{lightgraybg}\multicolumn{6}{c}{\textit{ LLaVA-NeXT-7B (Mistral)}}\\
    \textcolor{textgray}{Vanilla}&\textcolor{textgray}{73:44}&\textcolor{textgray}{391.9}&\textcolor{textgray}{391.0}&\textcolor{textgray}{495.1}&\textcolor{textgray}{86.8}  \\
    % 4424
    \midrule
    FastV&44:21 \textcolor{textgreen}{($\downarrow39.8\%$)}&202.7 \textcolor{textgreen}{($\downarrow48.2\%$)}&201.9 \textcolor{textgreen}{($\downarrow48.3\%$)}&249.1 \textcolor{textgreen}{($\downarrow49.6\%$)}&85.5 \textcolor{textgreen}{($\downarrow1.5\%$)}\\ 
    DART&46:17 \textcolor{textgreen}{($\downarrow37.2\%$)}&205.4 \textcolor{textgreen}{($\downarrow47.5\%$)}&205.2 \textcolor{textgreen}{($\downarrow47.5\%$)}&277.4 \textcolor{textgreen}{($\downarrow43.9\%$)}&86.0 \textcolor{textgreen}{($\downarrow0.9\%$)}\\
    ShortV&59:48 \textcolor{textgreen}{($\downarrow18.8\%$)}&284.7 \textcolor{textgreen}{($\downarrow27.3\%$)}&284.1 \textcolor{textgreen}{($\downarrow27.3\%$)}&382.4 \textcolor{textgreen}{($\downarrow22.7\%$)}&86.5 \textcolor{textgreen}{($\downarrow0.3\%$)}\\
    \rowcolor{lightredbg}\textbf{HalfV (Ours)}&\textbf{45:03} \textcolor{textgreen}{($\downarrow38.9\%$)}&\textbf{201.8} \textcolor{textgreen}{($\downarrow48.5\%$)}&\textbf{201.1} \textcolor{textgreen}{($\downarrow48.5\%$)}&\textbf{263.2} \textcolor{textgreen}{($\downarrow46.8\%$)}&\textbf{87.3} \textcolor{textred}{($\uparrow0.6\%$)}\\
    \midrule
    \rowcolor{lightgraybg}\multicolumn{6}{c}{\textit{Qwen25-VL-7B}}\\
    %2996
    \textcolor{textgray}{Vanilla}&\textcolor{textgray}{49:56}&\textcolor{textgray}{171.6}&-&\textcolor{textgray}{200.5}&\textcolor{textgray}{86.5}  \\
    \midrule
    FastV&25:03 \textcolor{textgreen}{($\downarrow49.8\%$)}&137.2 \textcolor{textgreen}{($\downarrow20.0\%$)}&-&165.7 \textcolor{textgreen}{($\downarrow17.3\%$)}&80.7 \textcolor{textgreen}{($\downarrow6.7\%$)}\\
    DART&26:44 \textcolor{textgreen}{($\downarrow46.5\%$)}&140.5 \textcolor{textgreen}{($\downarrow18.1\%$)}&-&169.9 \textcolor{textgreen}{($\downarrow15.3\%$)}&82.9 \textcolor{textgreen}{($\downarrow4.2\%$)}\\
    
    \rowcolor{lightredbg}\textbf{HalfV (Ours)}&\textbf{25:41} \textcolor{textgreen}{($\downarrow48.6\%$)}&\textbf{139.4} \textcolor{textgreen}{($\downarrow18.8\%$)}&-&\textbf{168.4} \textcolor{textgreen}{($\downarrow16.0\%$)}&\textbf{84.5} \textcolor{textgreen}{($\downarrow2.3\%$)}\\
    \bottomrule
    \end{tabular}
    }
    \caption{\textbf{Evaluation of efficiency for different models on POPE.} Inference costs of Total Time, Time To First Token (TTFT), Prefill Time, Latency on POPE dataset for LLaVA-NeXT-7B and Qwen25-VL-7B models.}
    \label{tab:efficiency}
    \vspace{-0.2cm}
\end{table}
\begin{table}[t]
    \centering
    \resizebox{\columnwidth}{!}{
    \begin{tabular}{cccccccc}
    \toprule
\textbf{\#}&\textbf{IVR}&\textbf{SSR}&\textbf{Perf.}&\textbf{TTFT}&\textbf{Speedup}&\textbf{MO}&\textbf{MO/TTFT}\\
    \midrule
    \rowcolor{lightgraybg}\multicolumn{8}{c}{\textit{ LLaVA-NeXT-7B (Mistral)}}\\ 
    % 418 375
         (a)&\CheckmarkBold& &67.1 \textcolor{textgreen}{($\downarrow0.3\%$)}&280.9 ms&$1.48\times$&20.6 ms&7.3\% \\
         (b)&&\CheckmarkBold &67.3 \textcolor{textgreen}{($\downarrow0.0\%$)}&332.7 ms&$1.26\times$&0.8 ms&0.2\% \\
         (c)&\CheckmarkBold&\CheckmarkBold &67.0 \textcolor{textgreen}{($\downarrow0.4\%$)}&220.9 ms&$1.89\times$&21.7 ms&9.8\% \\
         \midrule
    \rowcolor{lightgraybg}\multicolumn{8}{c}{\textit{ Qwen25-VL-7B (Qwen2.5)}}\\
         (d)&\CheckmarkBold& &79.5 \textcolor{textgreen}{($\downarrow2.3\%$)}&173.0 ms&$1.84\times$&10.2 ms&5.8\% \\
         (e)&&\CheckmarkBold &81.4 \textcolor{textgreen}{($\downarrow0.0\%$)}&192.7 ms&$1.66\times$&1.4 ms&0.7\%\\
         (f)&\CheckmarkBold&\CheckmarkBold &79.7 \textcolor{textgreen}{($\downarrow2.0\%$)}&166.1 ms&$1.92\times$&10.6 ms&6.3\% \\
    \bottomrule
    \end{tabular}
    }
    \caption{\textbf{Ablation study of module efficiency on AI2D dataset.} \textbf{IVR} denotes the module that mitigates \emph{Intrinsic Visual Redundancy}; \textbf{SSR} denotes the module that suppresses \emph{Semantic Saturation Redundancy}. \textbf{MO} abbreviates \emph{module overhead}, and \textbf{MO/TTFT} measures the module’s relative time overhead as a fraction of TTFT.}
    \label{tab:albation_module}
    \vspace{-0.4cm}
\end{table}
% 为评估 HalfV 的真实加速效果，我们在 LLaVA-NeXT-7B 与 Qwen2.5-VL-7B 上，比较了多种方法在 POPE 任务中的平均 总推理时间（Total Time）、首 token 生成时间（TTFT）、预填充时间（Prefill）、端到端延迟以及准确率。我们沿用主表中的相同实验设置，对 FastV、DART、ShortV 以及 HalfV 进行测试。

% 如图 4 所示，在 LLaVA-NeXT-7B 上，HalfV 在将总推理时间降低 38.9%、TTFT 降低 48.5%、端到端延迟降低 46.8% 的同时，准确率不仅未下降，反而提升 0.6%。与其他方法相比，HalfV 能在相近的加速幅度下更好地保持原始性能。对于 Qwen2.5-VL-7B，HalfV 将总推理时间降低 48.6%，准确率仅下降 2.3%；相较于 DART，准确率提升 1.9%。

% 需要说明的是，我们对 FastV 的实现进行了优化。原始 FastV 需要显式获取注意力分数，这使其难以适配多种底层加速算子（如 FlashAttention 与 SDPA）。为公平比较在 SDPA 下各方法的真实延迟，我们在测试 FastV 时将其显式注意力计算替换为与我们方法一致的线性点积计算（见 Section 3.3）。因此，表 4 中 FastV 的时间开销相较原始实现更低。
To assess the practical speedup of HalfV, we benchmark multiple methods on the POPE task using LLaVA-NeXT-7B and Qwen2.5-VL-7B. We report the average total time, time-to-first-token (TTFT), prefill time, end-to-end latency, and performance. Following the same experimental settings as in the main table \ref{tab:main_results}, we evaluate FastV \cite{fastv}, DART \cite{dart}, ShortV \cite{shortv}, and our method. As shown in Table \ref{tab:efficiency}, on LLaVA-NeXT-7B, HalfV reduces Total Time by \textbf{38.9\%}, TTFT by \textbf{48.5\%}, and latency by \textbf{46.8\%}, while accuracy not only remains intact but also improves by \textbf{0.6\%}. Compared with prior methods, HalfV better preserves the original model performance under comparable speedups. On Qwen2.5-VL-7B, HalfV achieves a \textbf{48.6\%} reduction in Total Time with only a \textbf{2.3\%} accuracy drop, and improves accuracy by \textbf{1.9\%} over DART.

% We optimize FastV for a fair latency comparison. As the original FastV explicitly extracts attention scores and is incompatible with kernels such as FlashAttention and SDPA, we use the same linear dot-product formulation as ours under SDPA (see Section~\ref{sec:halfv}). Consequently, the FastV time in Table~\ref{tab:efficiency} is lower than that of its original implementation.

\section{Analysis and Discussion}
\subsection{Ablation Study for Module Efficiency}
% 为更清晰地分析各模块的贡献，我们对 HalfV 的两步加速策略进行了模块消融实验。作为一种两阶段方法，HalfV 的第一步旨在缓解来自 ViT 端的 IVR，第二步则进一步抑制由 LLM 内部第二阶段全局聚合引发的 SSR。我们分别在 LLaVA-NeXT-7B 与 Qwen2.5-VL-7B 上对这两个模块进行消融，并统计不同配置下的性能、TTFT、加速比以及模块开销（Module overhead）。其中，表 5 的 MO/TTFT 表示该模块的处理开销占 TTFT 的比例。
% 如表 5 所示，IVR 模块的开销在 LLaVA-NeXT-7B 上为 7.3\%，在 Qwen2.5-VL-7B 上为 5.8\%；而针对 SSR 的架构感知模块在两类模型中的开销均 低于 1\%。值得注意的是，在 Qwen2.5-VL-7B 上，相比仅处理 IVR 的配置（表 5(d)），同时处理 IVR 与 SSR 的配置（表 5(f)）不仅带来更高的加速比（1.92× vs 1.84×），性能还提升 0.3\%。这表明两类冗余的级联抑制并非简单叠加，而是从模型内部机制出发形成了相互增益的协同效果。
To quantify each module’s contribution, we perform ablations on HalfV’s two-step pipeline. The first step mitigates IVR from the ViT encoder, and the second step suppresses SSR caused by global aggregation in the LLM. We evaluate variants on LLaVA-NeXT-7B and Qwen2.5-VL-7B, reporting performance, TTFT, speedup, and module overhead (MO).
As shown in Table~\ref{tab:albation_module}, the IVR module accounts for \textbf{7.3\%} of TTFT on LLaVA-NeXT-7B and \textbf{5.8\%} on Qwen2.5-VL-7B. In contrast, the architecture-aware SSR module costs less than \textbf{1\%} on both models. Notably, on Qwen2.5-VL-7B, enabling both modules yields higher speedup than IVR-only and even improves performance by \textbf{0.3\%} (Table~\ref{tab:albation_module}(f) and \ref{tab:albation_module}(d)). This indicates that jointly mitigating the two types of redundancy is not a naive concatenation, but a complementary design that yields synergistic benefits from an internal model perspective.
\begin{figure}[!htbp]
    \centering
    \includegraphics[width=\linewidth]{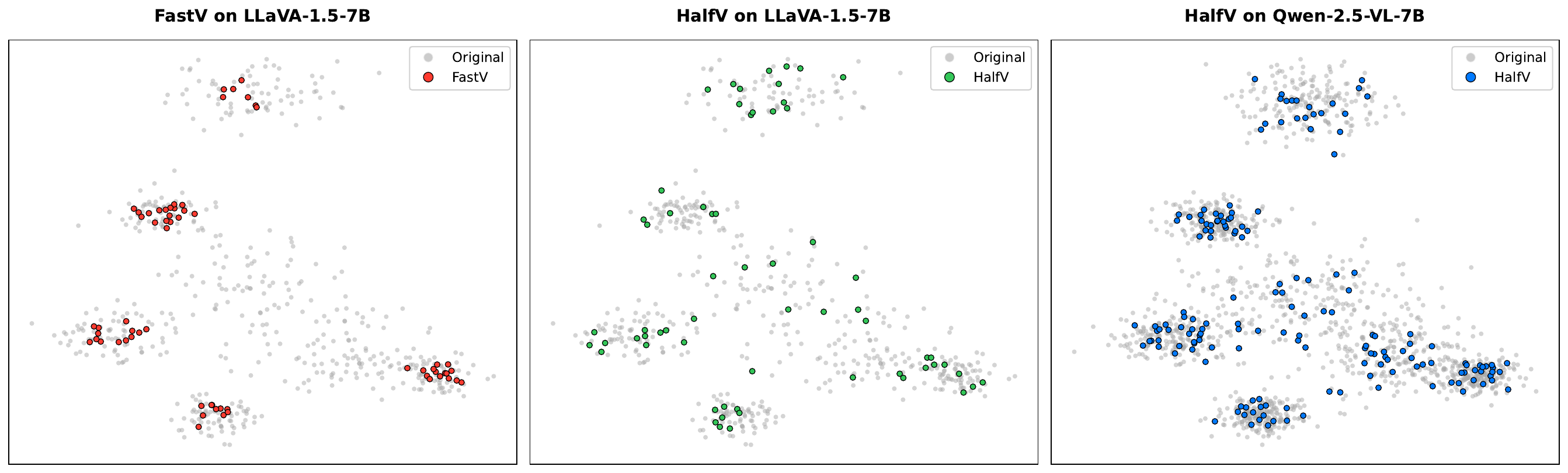}
    \vspace{-0.6cm}
    \caption{Comparison of feature space coverage}
    \vspace{-0.4cm}
    \label{fig:t-SNE}
\end{figure}
\subsection{Sensitivity Analysis of the Anchor Retention Ratio $R_{\mathcal{S}}$}
% 在HalfV的第一步裁剪过程中，我们首先需要选择锚点token。在小节3.2中，我们将其数量定义为$K_{\mathcal{S}}$。而在实际实现中，尤其是对于动态分辨率的模型，我们通过固定锚点token的保留率$R_{\mathcal{S}}$来确定其保留的数量。为了分析锚点token保留率对于模型性能的影响，我们在LLaVA-NeXT-7B和Qwen25-VL-7B模型中，仅使用第一步的裁剪策略，在不同层、不同锚点token保留率下进行了实验。具体来说，我们将全部裁剪比例固定为75\%，分别在LLaVA模型的第2、8、16层和Qwen模型的第2、8、21层进行第一步的裁剪策略。分别选取16和21层作为两个模型的分析层，是因为它们分别是这两个模型内部第三阶段的起始层(详细讨论见第3章节)。实验结果如图6所示，我们可以观察到在第2层时，选取较少的anchor token会较大地提高模型的最终性能，这说明模型在早期层需要依赖多样性的token获取更多特征的信息。与此同时，在第16/21层时，全部使用anchor token也可以实现接近100\%的性能。这可以间接说明模型在第二阶段聚合后，注意力机制可以很好地捕捉到模型内部的重要视觉信息，这也为HalfV第二部分的策略设计提供了数据上的支持。
In practice, we implement the anchor count as a ratio $R_\mathcal{S}$ to accommodate dynamic resolutions. We evaluate $R_{\mathcal{S}}$ on LLaVA-NeXT-7B and Qwen2.5-VL-7B, applying the first-step pruner at layers 2, 8, and the Stage III onset (Layer 16/21). As shown in Figure \ref{fig:anchor_ratio}, early layers (e.g., Layer 2) benefit from lower $R_{\mathcal{S}}$, indicating a reliance on token diversity. Conversely, deep layers maintain performance even with high anchor ratios, confirming that Stage II aggregation effectively concentrates key visual evidence, rendering extensive spatial coverage unnecessary in later stages.
We further visualize the selected visual tokens in the IVR stage on LLaVA-1.5v-13B and Qwen2.5-VL-7B. We apply t-SNE \citep{tsne} to project the $D$-dimensional visual tokens into a 2D space. As shown in Fig.~\ref{fig:t-SNE}, HalfV covers different clusters more uniformly, reducing the risk of spatial collapse.
\begin{figure}[!htbp]
    \centering
    % 第1张图
    \begin{subfigure}[b]{0.45\columnwidth}
        \centering
        \includegraphics[width=\textwidth]{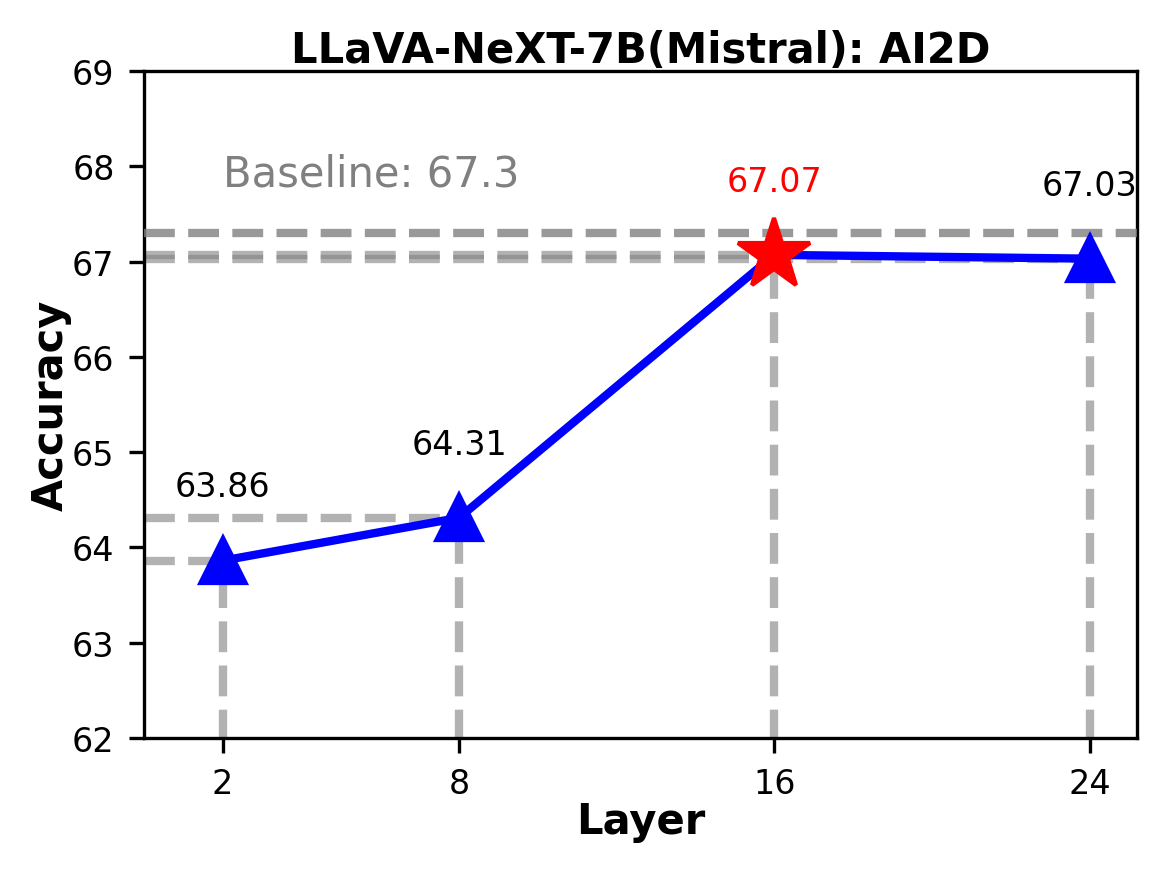}
        \label{fig:sub1}
    \end{subfigure}
    \hfill % 弹性空格，把图片撑开
    % 第2张图
    \begin{subfigure}[b]{0.45\columnwidth}
        \centering
        \includegraphics[width=\textwidth]{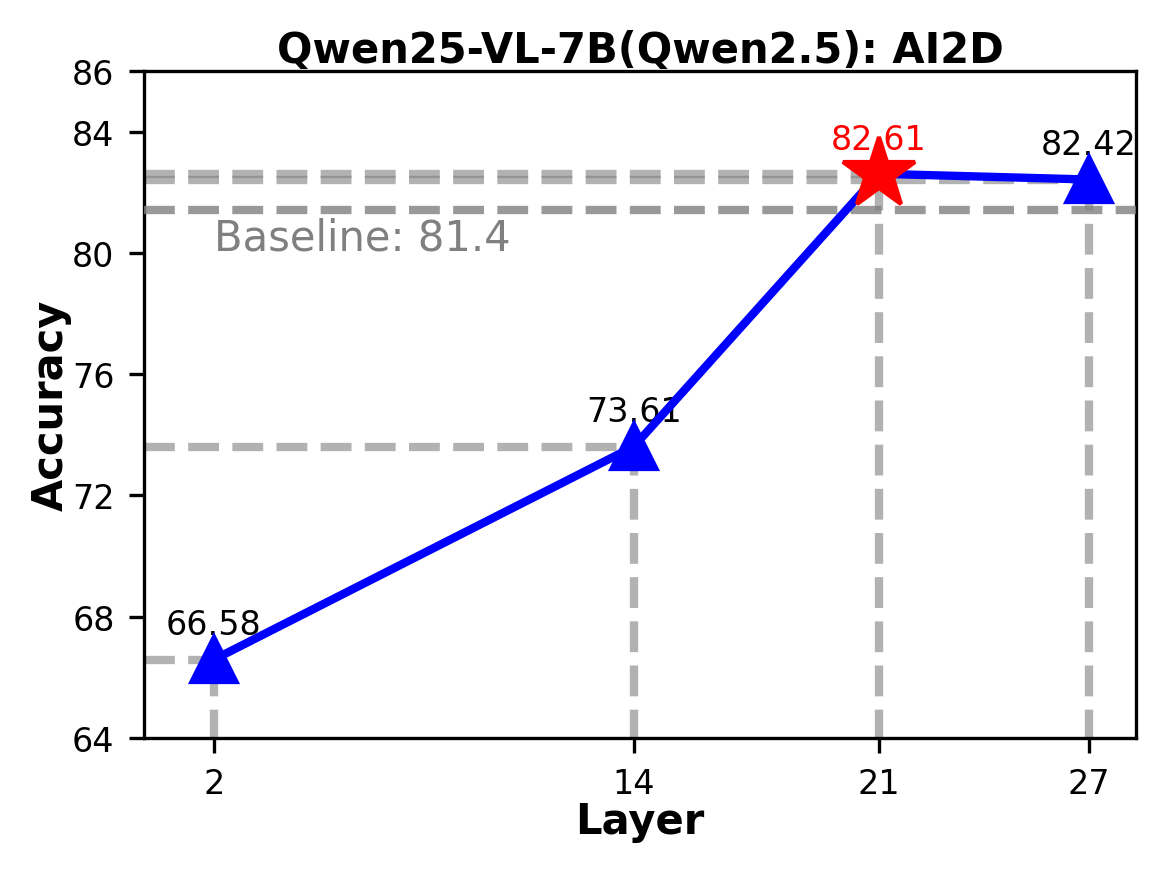}
        \label{fig:sub2}
    \end{subfigure}
    \vspace{-0.6cm}
    \caption{Impact of SSR start layer on different models}
    \vspace{-0.4cm}
    \label{fig:layer_ablation}
\end{figure}
\subsection{Sensitivity Analysis of the SSR Start Layer}
We evaluate the SSR module in isolation on LLaVA-NeXT-7B and Qwen2.5-VL-7B across varying start layers: Layer 2 (IVR), mid-Stage II (8/14), Stage III onset (16/21), and deep Stage III (24/27). As shown in Figure \ref{fig:layer_ablation}, early application significantly degrades performance. Conversely, initiating SSR at the Stage III onset yields near-lossless results, even surpassing the Qwen baseline. Notably, delaying SSR further into Stage III also harms performance. This confirms that the optimal insertion point is not "the later the better," but rather the Stage II-to-III transition (the information aggregation “valley”).
\section{Conclusion}
This paper tackles the limited cross-architecture generalization of existing MLLM acceleration methods and proposes an acceleration framework, HalfV. Using truncated matrix entropy as a unified probe, it reveals a three-stage redundancy evolution during inference and decomposes visual redundancy into Intrinsic Visual Redundancy (IVR) and Secondary Saturation Redundancy (SSR). HalfV then applies a two-step acceleration method to improve efficiency across model backbones.
\section{Limitations}
Although HalfV demonstrates strong and consistent improvements across multiple MLLMs and benchmarks, its applicability is constrained by practical considerations. First, similar to many existing inference acceleration methods, HalfV requires access to token-level hidden states during inference, and thus cannot be directly applied to closed-source black-box models such as the GPT and Gemini families. Second, due to current compute and hardware limitations, we have only deployed and evaluated HalfV on MLLMs in the 4B--13B parameter range, and we have not yet conducted systematic experiments on larger-scale models. In addition, to balance completeness and readability under the page limit, we defer some implementation details of evaluation metrics as well as further analyses and observations about internal model behaviors to the appendix. The main text presents the method, core experiments, and primary findings in full, ensuring that the key contributions and conclusions can be understood without relying on the appendix.
\bibliography{custom}

@article{internvl,
  author       = {Zhe Chen and
                  Jiannan Wu and
                  Wenhai Wang and
                  Weijie Su and
                  Guo Chen and
                  Sen Xing and
                  Muyan Zhong and
                  Qinglong Zhang and
                  Xizhou Zhu and
                  Lewei Lu and
                  Bin Li and
                  Ping Luo and
                  Tong Lu and
                  Yu Qiao and
                  Jifeng Dai},
  title        = {InternVL: Scaling up Vision Foundation Models and Aligning for Generic
                  Visual-Linguistic Tasks},
  journal      = {CoRR},
  volume       = {abs/2312.14238},
  year         = {2023},
  url          = {https://doi.org/10.48550/arXiv.2312.14238},
  doi          = {10.48550/ARXIV.2312.14238},
  eprinttype    = {arXiv},
  eprint       = {2312.14238},
  bibsource    = {dblp computer science bibliography, https://dblp.org}
}

@article{qwen2vl,
  author       = {Peng Wang and
                  Shuai Bai and
                  Sinan Tan and
                  Shijie Wang and
                  Zhihao Fan and
                  Jinze Bai and
                  Keqin Chen and
                  Xuejing Liu and
                  Jialin Wang and
                  Wenbin Ge and
                  Yang Fan and
                  Kai Dang and
                  Mengfei Du and
                  Xuancheng Ren and
                  Rui Men and
                  Dayiheng Liu and
                  Chang Zhou and
                  Jingren Zhou and
                  Junyang Lin},
  title        = {Qwen2-VL: Enhancing Vision-Language Model's Perception of the
                  World at Any Resolution},
  journal      = {CoRR},
  volume       = {abs/2409.12191},
  year         = {2024},
  url          = {https://doi.org/10.48550/arXiv.2409.12191},
  doi          = {10.48550/ARXIV.2409.12191},
  eprinttype    = {arXiv},
  eprint       = {2409.12191},
  bibsource    = {dblp computer science bibliography, https://dblp.org}
}

@article{qwen25vl,
  author       = {Shuai Bai and
                  Keqin Chen and
                  Xuejing Liu and
                  Jialin Wang and
                  Wenbin Ge and
                  Sibo Song and
                  Kai Dang and
                  Peng Wang and
                  Shijie Wang and
                  Jun Tang and
                  Humen Zhong and
                  Yuanzhi Zhu and
                  Ming{-}Hsuan Yang and
                  Zhaohai Li and
                  Jianqiang Wan and
                  Pengfei Wang and
                  Wei Ding and
                  Zheren Fu and
                  Yiheng Xu and
                  Jiabo Ye and
                  Xi Zhang and
                  Tianbao Xie and
                  Zesen Cheng and
                  Hang Zhang and
                  Zhibo Yang and
                  Haiyang Xu and
                  Junyang Lin},
  title        = {Qwen2.5-VL Technical Report},
  journal      = {CoRR},
  volume       = {abs/2502.13923},
  year         = {2025},
  url          = {https://doi.org/10.48550/arXiv.2502.13923},
  doi          = {10.48550/ARXIV.2502.13923},
  eprinttype    = {arXiv},
  eprint       = {2502.13923},
  bibsource    = {dblp computer science bibliography, https://dblp.org}
}

@inproceedings{llava-vl,
  author       = {Haotian Liu and
                  Chunyuan Li and
                  Yuheng Li and
                  Yong Jae Lee},
  title        = {Improved Baselines with Visual Instruction Tuning},
  booktitle    = {{IEEE/CVF} Conference on Computer Vision and Pattern Recognition,
                  {CVPR} 2024, Seattle, WA, USA, June 16-22, 2024},
  pages        = {26286--26296},
  publisher    = {{IEEE}},
  year         = {2024},
  url          = {https://doi.org/10.1109/CVPR52733.2024.02484},
  doi          = {10.1109/CVPR52733.2024.02484},
  bibsource    = {dblp computer science bibliography, https://dblp.org}
}

@inproceedings{llava,
  author       = {Haotian Liu and
                  Chunyuan Li and
                  Qingyang Wu and
                  Yong Jae Lee},
  editor       = {Alice Oh and
                  Tristan Naumann and
                  Amir Globerson and
                  Kate Saenko and
                  Moritz Hardt and
                  Sergey Levine},
  title        = {Visual Instruction Tuning},
  booktitle    = {Advances in Neural Information Processing Systems 36: Annual Conference
                  on Neural Information Processing Systems 2023, NeurIPS 2023, New Orleans,
                  LA, USA, December 10 - 16, 2023},
  year         = {2023},
  url          = {http://papers.nips.cc/paper\_files/paper/2023/hash/6dcf277ea32ce3288914faf369fe6de0-Abstract-Conference.html},
  bibsource    = {dblp computer science bibliography, https://dblp.org}
}

@inproceedings{fastv,
  author       = {Liang Chen and
                  Haozhe Zhao and
                  Tianyu Liu and
                  Shuai Bai and
                  Junyang Lin and
                  Chang Zhou and
                  Baobao Chang},
  editor       = {Ales Leonardis and
                  Elisa Ricci and
                  Stefan Roth and
                  Olga Russakovsky and
                  Torsten Sattler and
                  G{\"{u}}l Varol},
  title        = {An Image is Worth 1/2 Tokens After Layer 2: Plug-and-Play Inference
                  Acceleration for Large Vision-Language Models},
  booktitle    = {Computer Vision - {ECCV} 2024 - 18th European Conference, Milan, Italy,
                  September 29-October 4, 2024, Proceedings, Part {LXXXI}},
  series       = {Lecture Notes in Computer Science},
  volume       = {15139},
  pages        = {19--35},
  publisher    = {Springer},
  year         = {2024},
  url          = {https://doi.org/10.1007/978-3-031-73004-7\_2},
  doi          = {10.1007/978-3-031-73004-7\_2},
  bibsource    = {dblp computer science bibliography, https://dblp.org}
}

@inproceedings{topv,
  author       = {Cheng Yang and
                  Yang Sui and
                  Jinqi Xiao and
                  Lingyi Huang and
                  Yu Gong and
                  Chendi Li and
                  Jinghua Yan and
                  Yu Bai and
                  Ponnuswamy Sadayappan and
                  Xia Hu and
                  Bo Yuan},
  title        = {TopV: Compatible Token Pruning with Inference Time Optimization for
                  Fast and Low-Memory Multimodal Vision Language Model},
  booktitle    = {{IEEE/CVF} Conference on Computer Vision and Pattern Recognition,
                  {CVPR} 2025, Nashville, TN, USA, June 11-15, 2025},
  pages        = {19803--19813},
  publisher    = {Computer Vision Foundation / {IEEE}},
  year         = {2025},
  url          = {https://openaccess.thecvf.com/content/CVPR2025/html/Yang\_TopV\_Compatible\_Token\_Pruning\_with\_Inference\_Time\_Optimization\_for\_Fast\_CVPR\_2025\_paper.html},
  doi          = {10.1109/CVPR52734.2025.01844},
  bibsource    = {dblp computer science bibliography, https://dblp.org}
}

@article{sparsevlm,
  author       = {Yuan Zhang and
                  Chun{-}Kai Fan and
                  Junpeng Ma and
                  Wenzhao Zheng and
                  Tao Huang and
                  Kuan Cheng and
                  Denis A. Gudovskiy and
                  Tomoyuki Okuno and
                  Yohei Nakata and
                  Kurt Keutzer and
                  Shanghang Zhang},
  title        = {SparseVLM: Visual Token Sparsification for Efficient Vision-Language
                  Model Inference},
  journal      = {CoRR},
  volume       = {abs/2410.04417},
  year         = {2024},
  url          = {https://doi.org/10.48550/arXiv.2410.04417},
  doi          = {10.48550/ARXIV.2410.04417},
  eprinttype    = {arXiv},
  eprint       = {2410.04417},
  bibsource    = {dblp computer science bibliography, https://dblp.org}
}

@inproceedings{HiRED,
  author       = {Kazi Hasan Ibn Arif and
                  JinYi Yoon and
                  Dimitrios S. Nikolopoulos and
                  Hans Vandierendonck and
                  Deepu John and
                  Bo Ji},
  editor       = {Toby Walsh and
                  Julie Shah and
                  Zico Kolter},
  title        = {HiRED: Attention-Guided Token Dropping for Efficient Inference of
                  High-Resolution Vision-Language Models},
  booktitle    = {AAAI-25, Sponsored by the Association for the Advancement of Artificial
                  Intelligence, February 25 - March 4, 2025, Philadelphia, PA, {USA}},
  pages        = {1773--1781},
  publisher    = {{AAAI} Press},
  year         = {2025},
  url          = {https://doi.org/10.1609/aaai.v39i2.32171},
  doi          = {10.1609/AAAI.V39I2.32171},
  bibsource    = {dblp computer science bibliography, https://dblp.org}
}

@inproceedings{fitprune,
  author       = {Weihao Ye and
                  Qiong Wu and
                  Wenhao Lin and
                  Yiyi Zhou},
  editor       = {Toby Walsh and
                  Julie Shah and
                  Zico Kolter},
  title        = {Fit and Prune: Fast and Training-free Visual Token Pruning for Multi-modal
                  Large Language Models},
  booktitle    = {AAAI-25, Sponsored by the Association for the Advancement of Artificial
                  Intelligence, February 25 - March 4, 2025, Philadelphia, PA, {USA}},
  pages        = {22128--22136},
  publisher    = {{AAAI} Press},
  year         = {2025},
  url          = {https://doi.org/10.1609/aaai.v39i21.34366},
  doi          = {10.1609/AAAI.V39I21.34366},
  bibsource    = {dblp computer science bibliography, https://dblp.org}
}

@inproceedings{vit,
  author       = {Alexey Dosovitskiy and
                  Lucas Beyer and
                  Alexander Kolesnikov and
                  Dirk Weissenborn and
                  Xiaohua Zhai and
                  Thomas Unterthiner and
                  Mostafa Dehghani and
                  Matthias Minderer and
                  Georg Heigold and
                  Sylvain Gelly and
                  Jakob Uszkoreit and
                  Neil Houlsby},
  title        = {An Image is Worth 16x16 Words: Transformers for Image Recognition
                  at Scale},
  booktitle    = {9th International Conference on Learning Representations, {ICLR} 2021,
                  Virtual Event, Austria, May 3-7, 2021},
  publisher    = {OpenReview.net},
  year         = {2021},
  url          = {https://openreview.net/forum?id=YicbFdNTTy},
  bibsource    = {dblp computer science bibliography, https://dblp.org}
}

@article{llava-prumerge,
  author       = {Yuzhang Shang and
                  Mu Cai and
                  Bingxin Xu and
                  Yong Jae Lee and
                  Yan Yan},
  title        = {LLaVA-PruMerge: Adaptive Token Reduction for Efficient Large Multimodal
                  Models},
  journal      = {CoRR},
  volume       = {abs/2403.15388},
  year         = {2024},
  url          = {https://doi.org/10.48550/arXiv.2403.15388},
  doi          = {10.48550/ARXIV.2403.15388},
  eprinttype    = {arXiv},
  eprint       = {2403.15388},
  bibsource    = {dblp computer science bibliography, https://dblp.org}
}

@article{pdrop,
  author       = {Long Xing and
                  Qidong Huang and
                  Xiaoyi Dong and
                  Jiajie Lu and
                  Pan Zhang and
                  Yuhang Zang and
                  Yuhang Cao and
                  Conghui He and
                  Jiaqi Wang and
                  Feng Wu and
                  Dahua Lin},
  title        = {PyramidDrop: Accelerating Your Large Vision-Language Models via Pyramid
                  Visual Redundancy Reduction},
  journal      = {CoRR},
  volume       = {abs/2410.17247},
  year         = {2024},
  url          = {https://doi.org/10.48550/arXiv.2410.17247},
  doi          = {10.48550/ARXIV.2410.17247},
  eprinttype    = {arXiv},
  eprint       = {2410.17247},
  bibsource    = {dblp computer science bibliography, https://dblp.org}
}

@misc{feather,
      title={Feather the Throttle: Revisiting Visual Token Pruning for Vision-Language Model Acceleration}, 
      author={Mark Endo and Xiaohan Wang and Serena Yeung-Levy},
      year={2025},
      eprint={2412.13180},
      archivePrefix={arXiv},
      primaryClass={cs.CV},
      url={https://arxiv.org/abs/2412.13180}, 
}

@article{dart,
  author       = {Zichen Wen and
                  Yifeng Gao and
                  Shaobo Wang and
                  Junyuan Zhang and
                  Qintong Zhang and
                  Weijia Li and
                  Conghui He and
                  Linfeng Zhang},
  title        = {Stop Looking for Important Tokens in Multimodal Language Models: Duplication
                  Matters More},
  journal      = {CoRR},
  volume       = {abs/2502.11494},
  year         = {2025},
  url          = {https://doi.org/10.48550/arXiv.2502.11494},
  doi          = {10.48550/ARXIV.2502.11494},
  eprinttype    = {arXiv},
  eprint       = {2502.11494},
  bibsource    = {dblp computer science bibliography, https://dblp.org}
}

@inproceedings{divprune,
  author       = {Saeed Ranjbar Alvar and
                  Gursimran Singh and
                  Mohammad Akbari and
                  Yong Zhang},
  title        = {DivPrune: Diversity-based Visual Token Pruning for Large Multimodal
                  Models},
  booktitle    = {{IEEE/CVF} Conference on Computer Vision and Pattern Recognition,
                  {CVPR} 2025, Nashville, TN, USA, June 11-15, 2025},
  pages        = {9392--9401},
  publisher    = {Computer Vision Foundation / {IEEE}},
  year         = {2025},
  url          = {https://openaccess.thecvf.com/content/CVPR2025/html/Alvar\_DivPrune\_Diversity-based\_Visual\_Token\_Pruning\_for\_Large\_Multimodal\_Models\_CVPR\_2025\_paper.html},
  doi          = {10.1109/CVPR52734.2025.00877},
  bibsource    = {dblp computer science bibliography, https://dblp.org}
}

@article{holov,
  author       = {Xin Zou and
                  Di Lu and
                  Yizhou Wang and
                  Yibo Yan and
                  Yuanhuiyi Lyu and
                  Xu Zheng and
                  Linfeng Zhang and
                  Xuming Hu},
  title        = {Don't Just Chase "Highlighted Tokens" in MLLMs: Revisiting
                  Visual Holistic Context Retention},
  journal      = {CoRR},
  volume       = {abs/2510.02912},
  year         = {2025},
  url          = {https://doi.org/10.48550/arXiv.2510.02912},
  doi          = {10.48550/ARXIV.2510.02912},
  eprinttype    = {arXiv},
  eprint       = {2510.02912},
  bibsource    = {dblp computer science bibliography, https://dblp.org}
}

@article{flowcut,
  author       = {Jintao Tong and
                  Wenwei Jin and
                  Pengda Qin and
                  Anqi Li and
                  Yixiong Zou and
                  Yuhong Li and
                  Yuhua Li and
                  Ruixuan Li},
  title        = {FlowCut: Rethinking Redundancy via Information Flow for Efficient
                  Vision-Language Models},
  journal      = {CoRR},
  volume       = {abs/2505.19536},
  year         = {2025},
  url          = {https://doi.org/10.48550/arXiv.2505.19536},
  doi          = {10.48550/ARXIV.2505.19536},
  eprinttype    = {arXiv},
  eprint       = {2505.19536},
  bibsource    = {dblp computer science bibliography, https://dblp.org}
}

@article{shortv,
  author       = {Qianhao Yuan and
                  Qingyu Zhang and
                  Yanjiang Liu and
                  Jiawei Chen and
                  Yaojie Lu and
                  Hongyu Lin and
                  Jia Zheng and
                  Xianpei Han and
                  Le Sun},
  title        = {ShortV: Efficient Multimodal Large Language Models by Freezing Visual
                  Tokens in Ineffective Layers},
  journal      = {CoRR},
  volume       = {abs/2504.00502},
  year         = {2025},
  url          = {https://doi.org/10.48550/arXiv.2504.00502},
  doi          = {10.48550/ARXIV.2504.00502},
  eprinttype    = {arXiv},
  eprint       = {2504.00502},
  bibsource    = {dblp computer science bibliography, https://dblp.org}
}

@inproceedings{truncated,
  author       = {Yifan Zhang and
                  Zhiquan Tan and
                  Jingqin Yang and
                  Weiran Huang and
                  Yang Yuan},
  title        = {Matrix Information Theory for Self-Supervised Learning},
  booktitle    = {Forty-first International Conference on Machine Learning, {ICML} 2024,
                  Vienna, Austria, July 21-27, 2024},
  publisher    = {OpenReview.net},
  year         = {2024},
  url          = {https://openreview.net/forum?id=wleAlsklEh},
  bibsource    = {dblp computer science bibliography, https://dblp.org}
}

@inproceedings{uncomp,
    title = "{UNC}omp: Can Matrix Entropy Uncover Sparsity? {---} A Compressor Design from an Uncertainty-Aware Perspective",
    author = "Xiong, Jing  and
      Shen, Jianghan  and
      Ye, Fanghua  and
      Tao, Chaofan  and
      Wan, Zhongwei  and
      Lu, Jianqiao  and
      Wu, Xun  and
      Zheng, Chuanyang  and
      Guo, Zhijiang  and
      Yang, Min  and
      Kong, Lingpeng  and
      Wong, Ngai",
    editor = "Christodoulopoulos, Christos  and
      Chakraborty, Tanmoy  and
      Rose, Carolyn  and
      Peng, Violet",
    booktitle = "Proceedings of the 2025 Conference on Empirical Methods in Natural Language Processing",
    month = nov,
    year = "2025",
    address = "Suzhou, China",
    publisher = "Association for Computational Linguistics",
    url = "https://aclanthology.org/2025.emnlp-main.209/",
    doi = "10.18653/v1/2025.emnlp-main.209",
    pages = "4179--4199",
    ISBN = "979-8-89176-332-6"
}

@article{dyvte,
  author       = {Qiong Wu and
                  Wenhao Lin and
                  Weihao Ye and
                  Yiyi Zhou and
                  Xiaoshuai Sun and
                  Rongrong Ji},
  title        = {Accelerating Multimodal Large Language Models via Dynamic Visual-Token
                  Exit and the Empirical Findings},
  journal      = {CoRR},
  volume       = {abs/2411.19628},
  year         = {2024},
  url          = {https://doi.org/10.48550/arXiv.2411.19628},
  doi          = {10.48550/ARXIV.2411.19628},
  eprinttype    = {arXiv},
  eprint       = {2411.19628},
  bibsource    = {dblp computer science bibliography, https://dblp.org}
}

@inproceedings{vtw,
  author       = {Zhihang Lin and
                  Mingbao Lin and
                  Luxi Lin and
                  Rongrong Ji},
  editor       = {Toby Walsh and
                  Julie Shah and
                  Zico Kolter},
  title        = {Boosting Multimodal Large Language Models with Visual Tokens Withdrawal
                  for Rapid Inference},
  booktitle    = {AAAI-25, Sponsored by the Association for the Advancement of Artificial
                  Intelligence, February 25 - March 4, 2025, Philadelphia, PA, {USA}},
  pages        = {5334--5342},
  publisher    = {{AAAI} Press},
  year         = {2025},
  url          = {https://doi.org/10.1609/aaai.v39i5.32567},
  doi          = {10.1609/AAAI.V39I5.32567},
  bibsource    = {dblp computer science bibliography, https://dblp.org}
}

@article{btp,
  author       = {Kaiyuan Li and
                  Xiaoyue Chen and
                  Chen Gao and
                  Yong Li and
                  Xinlei Chen},
  title        = {Balanced Token Pruning: Accelerating Vision Language Models Beyond
                  Local Optimization},
  journal      = {CoRR},
  volume       = {abs/2505.22038},
  year         = {2025},
  url          = {https://doi.org/10.48550/arXiv.2505.22038},
  doi          = {10.48550/ARXIV.2505.22038},
  eprinttype    = {arXiv},
  eprint       = {2505.22038},
  bibsource    = {dblp computer science bibliography, https://dblp.org}
}

@inproceedings{rep,
  author       = {Nick Jiang and
                  Anish Kachinthaya and
                  Suzanne Petryk and
                  Yossi Gandelsman},
  title        = {Interpreting and Editing Vision-Language Representations to Mitigate
                  Hallucinations},
  booktitle    = {The Thirteenth International Conference on Learning Representations,
                  {ICLR} 2025, Singapore, April 24-28, 2025},
  publisher    = {OpenReview.net},
  year         = {2025},
  url          = {https://openreview.net/forum?id=94kQgWXojH},
  bibsource    = {dblp computer science bibliography, https://dblp.org}
}

@inproceedings{act,
  author       = {Shiqi Chen and
                  Miao Xiong and
                  Junteng Liu and
                  Zhengxuan Wu and
                  Teng Xiao and
                  Siyang Gao and
                  Junxian He},
  title        = {In-Context Sharpness as Alerts: An Inner Representation Perspective
                  for Hallucination Mitigation},
  booktitle    = {Forty-first International Conference on Machine Learning, {ICML} 2024,
                  Vienna, Austria, July 21-27, 2024},
  publisher    = {OpenReview.net},
  year         = {2024},
  url          = {https://openreview.net/forum?id=s3e8poX3kb},
  bibsource    = {dblp computer science bibliography, https://dblp.org}
}

@article{elbow,
author = {Cangelosi, Richard and Goriely, Alain},
year = {2007},
month = {02},
pages = {2},
title = {Component retention in principal component analysis with application to cDNA microarray data},
volume = {2},
journal = {Biology direct},
doi = {10.1186/1745-6150-2-2}
}

@article{rope,
  author       = {Jianlin Su and
                  Murtadha H. M. Ahmed and
                  Yu Lu and
                  Shengfeng Pan and
                  Wen Bo and
                  Yunfeng Liu},
  title        = {RoFormer: Enhanced transformer with Rotary Position Embedding},
  journal      = {Neurocomputing},
  volume       = {568},
  pages        = {127063},
  year         = {2024},
  url          = {https://doi.org/10.1016/j.neucom.2023.127063},
  doi          = {10.1016/J.NEUCOM.2023.127063},
  bibsource    = {dblp computer science bibliography, https://dblp.org}
}

@inproceedings{visionzip,
  author       = {Senqiao Yang and
                  Yukang Chen and
                  Zhuotao Tian and
                  Chengyao Wang and
                  Jingyao Li and
                  Bei Yu and
                  Jiaya Jia},
  title        = {VisionZip: Longer is Better but Not Necessary in Vision Language Models},
  booktitle    = {{IEEE/CVF} Conference on Computer Vision and Pattern Recognition,
                  {CVPR} 2025, Nashville, TN, USA, June 11-15, 2025},
  pages        = {19792--19802},
  publisher    = {Computer Vision Foundation / {IEEE}},
  year         = {2025},
  url          = {https://openaccess.thecvf.com/content/CVPR2025/html/Yang\_VisionZip\_Longer\_is\_Better\_but\_Not\_Necessary\_in\_Vision\_Language\_CVPR\_2025\_paper.html},
  doi          = {10.1109/CVPR52734.2025.01843},
  bibsource    = {dblp computer science bibliography, https://dblp.org}
}

@misc{llavanext,
    title={LLaVA-NeXT: Improved reasoning, OCR, and world knowledge},
    url={https://llava-vl.github.io/blog/2024-01-30-llava-next/},
    author={Liu, Haotian and Li, Chunyuan and Li, Yuheng and Li, Bo and Zhang, Yuanhan and Shen, Sheng and Lee, Yong Jae},
    month={January},
    year={2024}
}

@inproceedings{gqa,
  author       = {Drew A. Hudson and
                  Christopher D. Manning},
  title        = {{GQA:} {A} New Dataset for Real-World Visual Reasoning and Compositional
                  Question Answering},
  booktitle    = {{IEEE} Conference on Computer Vision and Pattern Recognition, {CVPR}
                  2019, Long Beach, CA, USA, June 16-20, 2019},
  pages        = {6700--6709},
  publisher    = {Computer Vision Foundation / {IEEE}},
  year         = {2019},
  url          = {http://openaccess.thecvf.com/content\_CVPR\_2019/html/Hudson\_GQA\_A\_New\_Dataset\_for\_Real-World\_Visual\_Reasoning\_and\_Compositional\_CVPR\_2019\_paper.html},
  doi          = {10.1109/CVPR.2019.00686},
  bibsource    = {dblp computer science bibliography, https://dblp.org}
}

@misc{mme,
      title={MME: A Comprehensive Evaluation Benchmark for Multimodal Large Language Models}, 
      author={Chaoyou Fu and Peixian Chen and Yunhang Shen and Yulei Qin and Mengdan Zhang and Xu Lin and Jinrui Yang and Xiawu Zheng and Ke Li and Xing Sun and Yunsheng Wu and Rongrong Ji and Caifeng Shan and Ran He},
      year={2025},
      eprint={2306.13394},
      archivePrefix={arXiv},
      primaryClass={cs.CV},
      url={https://arxiv.org/abs/2306.13394}, 
}

@inproceedings{pope,
  author       = {Yixiao He and
                  Haifeng Sun and
                  Pengfei Ren and
                  Jingyu Wang and
                  Huazheng Wang and
                  Qi Qi and
                  Zirui Zhuang and
                  Jing Wang},
  editor       = {Luis Chiruzzo and
                  Alan Ritter and
                  Lu Wang},
  title        = {Evaluating and Mitigating Object Hallucination in Large Vision-Language
                  Models: Can They Still See Removed Objects?},
  booktitle    = {Proceedings of the 2025 Conference of the Nations of the Americas
                  Chapter of the Association for Computational Linguistics: Human Language
                  Technologies, {NAACL} 2025 - Volume 1: Long Papers, Albuquerque, New
                  Mexico, USA, April 29 - May 4, 2025},
  pages        = {6841--6858},
  publisher    = {Association for Computational Linguistics},
  year         = {2025},
  url          = {https://doi.org/10.18653/v1/2025.naacl-long.349},
  doi          = {10.18653/V1/2025.NAACL-LONG.349},
  bibsource    = {dblp computer science bibliography, https://dblp.org}
}

@inproceedings{sqa,
  author       = {Pan Lu and
                  Swaroop Mishra and
                  Tanglin Xia and
                  Liang Qiu and
                  Kai{-}Wei Chang and
                  Song{-}Chun Zhu and
                  Oyvind Tafjord and
                  Peter Clark and
                  Ashwin Kalyan},
  editor       = {Sanmi Koyejo and
                  S. Mohamed and
                  A. Agarwal and
                  Danielle Belgrave and
                  K. Cho and
                  A. Oh},
  title        = {Learn to Explain: Multimodal Reasoning via Thought Chains for Science
                  Question Answering},
  booktitle    = {Advances in Neural Information Processing Systems 35: Annual Conference
                  on Neural Information Processing Systems 2022, NeurIPS 2022, New Orleans,
                  LA, USA, November 28 - December 9, 2022},
  year         = {2022},
  url          = {http://papers.nips.cc/paper\_files/paper/2022/hash/11332b6b6cf4485b84afadb1352d3a9a-Abstract-Conference.html},
  bibsource    = {dblp computer science bibliography, https://dblp.org}
}

@inproceedings{vizwiz,
  author       = {Jeffrey P. Bigham and
                  Chandrika Jayant and
                  Hanjie Ji and
                  Greg Little and
                  Andrew Miller and
                  Robert C. Miller and
                  Robin Miller and
                  Aubrey Tatarowicz and
                  Brandyn White and
                  Samuel White and
                  Tom Yeh},
  editor       = {Ken Perlin and
                  Mary Czerwinski and
                  Rob Miller},
  title        = {VizWiz: nearly real-time answers to visual questions},
  booktitle    = {Proceedings of the 23rd Annual {ACM} Symposium on User Interface Software
                  and Technology, New York, NY, USA, October 3-6, 2010},
  pages        = {333--342},
  publisher    = {{ACM}},
  year         = {2010},
  url          = {https://doi.org/10.1145/1866029.1866080},
  doi          = {10.1145/1866029.1866080},
  bibsource    = {dblp computer science bibliography, https://dblp.org}
}

@inproceedings{textvqa,
  author       = {Amanpreet Singh and
                  Vivek Natarajan and
                  Meet Shah and
                  Yu Jiang and
                  Xinlei Chen and
                  Dhruv Batra and
                  Devi Parikh and
                  Marcus Rohrbach},
  title        = {Towards {VQA} Models That Can Read},
  booktitle    = {{IEEE} Conference on Computer Vision and Pattern Recognition, {CVPR}
                  2019, Long Beach, CA, USA, June 16-20, 2019},
  pages        = {8317--8326},
  publisher    = {Computer Vision Foundation / {IEEE}},
  year         = {2019},
  url          = {http://openaccess.thecvf.com/content\_CVPR\_2019/html/Singh\_Towards\_VQA\_Models\_That\_Can\_Read\_CVPR\_2019\_paper.html},
  doi          = {10.1109/CVPR.2019.00851},
  bibsource    = {dblp computer science bibliography, https://dblp.org}
}

@inproceedings{mmb,
  author       = {Yuan Liu and
                  Haodong Duan and
                  Yuanhan Zhang and
                  Bo Li and
                  Songyang Zhang and
                  Wangbo Zhao and
                  Yike Yuan and
                  Jiaqi Wang and
                  Conghui He and
                  Ziwei Liu and
                  Kai Chen and
                  Dahua Lin},
  editor       = {Ales Leonardis and
                  Elisa Ricci and
                  Stefan Roth and
                  Olga Russakovsky and
                  Torsten Sattler and
                  G{\"{u}}l Varol},
  title        = {MMBench: Is Your Multi-modal Model an All-Around Player?},
  booktitle    = {Computer Vision - {ECCV} 2024 - 18th European Conference, Milan, Italy,
                  September 29-October 4, 2024, Proceedings, Part {VI}},
  series       = {Lecture Notes in Computer Science},
  volume       = {15064},
  pages        = {216--233},
  publisher    = {Springer},
  year         = {2024},
  url          = {https://doi.org/10.1007/978-3-031-72658-3\_13},
  doi          = {10.1007/978-3-031-72658-3\_13},
  bibsource    = {dblp computer science bibliography, https://dblp.org}
}

@inproceedings{ai2d,
  author       = {Aniruddha Kembhavi and
                  Mike Salvato and
                  Eric Kolve and
                  Min Joon Seo and
                  Hannaneh Hajishirzi and
                  Ali Farhadi},
  editor       = {Bastian Leibe and
                  Jiri Matas and
                  Nicu Sebe and
                  Max Welling},
  title        = {A Diagram is Worth a Dozen Images},
  booktitle    = {Computer Vision - {ECCV} 2016 - 14th European Conference, Amsterdam,
                  The Netherlands, October 11-14, 2016, Proceedings, Part {IV}},
  series       = {Lecture Notes in Computer Science},
  volume       = {9908},
  pages        = {235--251},
  publisher    = {Springer},
  year         = {2016},
  url          = {https://doi.org/10.1007/978-3-319-46493-0\_15},
  doi          = {10.1007/978-3-319-46493-0\_15},
  bibsource    = {dblp computer science bibliography, https://dblp.org}
}

@article{kl,
 ISSN = {00034851},
 URL = {http://www.jstor.org/stable/2236703},
 author = {S. Kullback and R. A. Leibler},
 journal = {The Annals of Mathematical Statistics},
 number = {1},
 pages = {79--86},
 publisher = {Institute of Mathematical Statistics},
 title = {On Information and Sufficiency},
 urldate = {2026-01-05},
 volume = {22},
 year = {1951}
}

@inproceedings{flashattention,
  author       = {Tri Dao and
                  Daniel Y. Fu and
                  Stefano Ermon and
                  Atri Rudra and
                  Christopher R{\'{e}}},
  editor       = {Sanmi Koyejo and
                  S. Mohamed and
                  A. Agarwal and
                  Danielle Belgrave and
                  K. Cho and
                  A. Oh},
  title        = {FlashAttention: Fast and Memory-Efficient Exact Attention with IO-Awareness},
  booktitle    = {Advances in Neural Information Processing Systems 35: Annual Conference
                  on Neural Information Processing Systems 2022, NeurIPS 2022, New Orleans,
                  LA, USA, November 28 - December 9, 2022},
  year         = {2022},
  url          = {http://papers.nips.cc/paper\_files/paper/2022/hash/67d57c32e20fd0a7a302cb81d36e40d5-Abstract-Conference.html},
  bibsource    = {dblp computer science bibliography, https://dblp.org}
}

@article{fps,
title = {Clustering to minimize the maximum intercluster distance},
journal = {Theoretical Computer Science},
volume = {38},
pages = {293-306},
year = {1985},
issn = {0304-3975},
doi = {https://doi.org/10.1016/0304-3975(85)90224-5},
url = {https://www.sciencedirect.com/science/article/pii/0304397585902245},
author = {Teofilo F. Gonzalez}
}

@misc{qwen3,
      title={Qwen3 Technical Report}, 
      author={An Yang and Anfeng Li and Baosong Yang and Beichen Zhang and Binyuan Hui and Bo Zheng and Bowen Yu and Chang Gao and Chengen Huang and Chenxu Lv and Chujie Zheng and Dayiheng Liu and Fan Zhou and Fei Huang and Feng Hu and Hao Ge and Haoran Wei and Huan Lin and Jialong Tang and Jian Yang and Jianhong Tu and Jianwei Zhang and Jianxin Yang and Jiaxi Yang and Jing Zhou and Jingren Zhou and Junyang Lin and Kai Dang and Keqin Bao and Kexin Yang and Le Yu and Lianghao Deng and Mei Li and Mingfeng Xue and Mingze Li and Pei Zhang and Peng Wang and Qin Zhu and Rui Men and Ruize Gao and Shixuan Liu and Shuang Luo and Tianhao Li and Tianyi Tang and Wenbiao Yin and Xingzhang Ren and Xinyu Wang and Xinyu Zhang and Xuancheng Ren and Yang Fan and Yang Su and Yichang Zhang and Yinger Zhang and Yu Wan and Yuqiong Liu and Zekun Wang and Zeyu Cui and Zhenru Zhang and Zhipeng Zhou and Zihan Qiu},
      year={2025},
      eprint={2505.09388},
      archivePrefix={arXiv},
      primaryClass={cs.CL},
      url={https://arxiv.org/abs/2505.09388}, 
}

@inproceedings{videollava,
  author       = {Bin Lin and
                  Yang Ye and
                  Bin Zhu and
                  Jiaxi Cui and
                  Munan Ning and
                  Peng Jin and
                  Li Yuan},
  editor       = {Yaser Al{-}Onaizan and
                  Mohit Bansal and
                  Yun{-}Nung Chen},
  title        = {Video-LLaVA: Learning United Visual Representation by Alignment Before
                  Projection},
  booktitle    = {Proceedings of the 2024 Conference on Empirical Methods in Natural
                  Language Processing, {EMNLP} 2024, Miami, FL, USA, November 12-16,
                  2024},
  pages        = {5971--5984},
  publisher    = {Association for Computational Linguistics},
  year         = {2024},
  url          = {https://doi.org/10.18653/v1/2024.emnlp-main.342},
  doi          = {10.18653/V1/2024.EMNLP-MAIN.342},
  bibsource    = {dblp computer science bibliography, https://dblp.org}
}

@article{tsne,
  author  = {Laurens van der Maaten and Geoffrey Hinton},
  title   = {Visualizing Data using t-SNE},
  journal = {Journal of Machine Learning Research},
  year    = {2008},
  volume  = {9},
  number  = {86},
  pages   = {2579--2605},
  url     = {http://jmlr.org/papers/v9/vandermaaten08a.html}
}

\appendix
\etocsettocdepth{subsection} % 重新允许写入（到 subsection）
\part*{Appendix}
\addcontentsline{toc}{part}{Appendix} % 可选：把“Appendix”写进目录
\etocsettocstyle{}{}   % ① 去掉 “Contents” 标题
\localtableofcontents 

\section{Implementation Details of HalfV}
\subsection{Layer-level Inactivity Implementation Details}
\label{appendix:layer_inactivaty}
\begin{figure}
    \centering
    \includegraphics[width=0.9\linewidth]{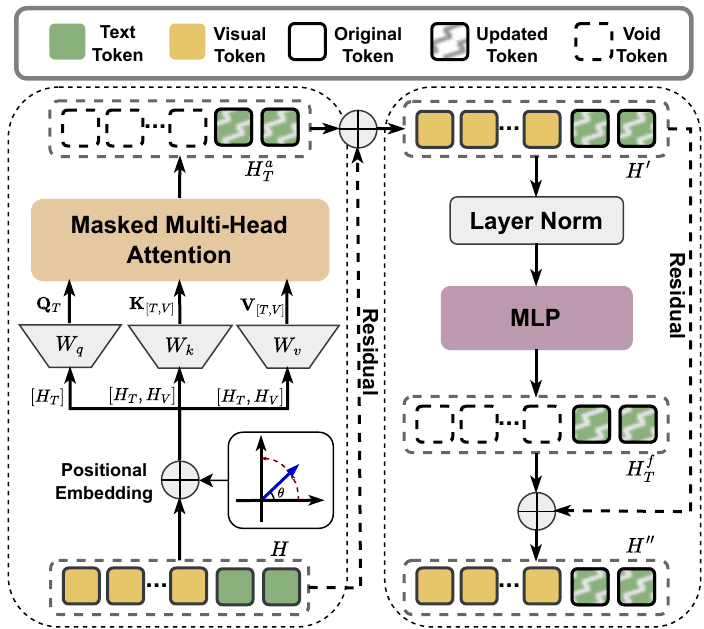}
    \caption{Illustration of Handling Inactivity Layers}
    \label{fig:inactivity}
\end{figure}
As shown in Figure~\ref{fig:inactivity}, we consider a decoder layer with input hidden states $\mathbf{H}\in\mathbb{R}^{N\times d}$,
where $N=N_V+N_T$ denotes the total number of tokens, consisting of $N_V$ visual tokens and $N_T$ textual tokens.
For notational convenience, we write the sequence as
$\mathbf{H}=\operatorname{Concat}(\mathbf{H}_V,\mathbf{H}_T)$ and similarly decompose the query matrix
$\mathbf{Q}=\operatorname{Concat}(\mathbf{Q}_V,\mathbf{Q}_T)$.
Keys and values are computed from all tokens, yielding $\mathbf{K},\mathbf{V}\in\mathbb{R}^{N\times d_k}$.

\paragraph{Self-Attention with Visual Update Termination.}
We only compute attention outputs for textual queries, while terminating the attention update for visual tokens. Specifically, the attention-induced updates are defined as
\begin{equation}
\Delta\mathbf{H}_T^{a}
=
\operatorname{softmax}\!\left(
\frac{\mathbf{Q}_T \mathbf{K}^{\top}}{\sqrt{d_k}}
\right)
\mathbf{V},
\quad
\Delta\mathbf{H}_V^{a}=\mathbf{0},
\label{eq:terminate_attn_update}
\end{equation}
where $\frac{\mathbf{Q}_T \mathbf{K}^{\top}}{\sqrt{d_k}}\in \mathbb{R}^{N_T\times N}$ and $d_k$ is the key dimension.
After the residual connection, the post-attention hidden states are
\begin{equation}
\mathbf{H}'=\mathbf{H}+\operatorname{Concat}(\Delta\mathbf{H}_V^{a},\Delta\mathbf{H}_T^{a}).
\label{eq:terminate_attn_residual}
\end{equation}
Note that this construction preserves visual states in the attention sublayer, i.e., $\mathbf{H}'_V=\mathbf{H}_V$.

\paragraph{FFN with Visual Update Termination.}
In the FFN sublayer, we similarly update only textual tokens and terminate updates for visual tokens:
\begin{equation}
\Delta\mathbf{H}_T^{f}=\operatorname{FFN}(\mathbf{H}'_T),
\qquad
\Delta\mathbf{H}_V^{f}=\mathbf{0}.
\label{eq:terminate_ffn_update}
\end{equation}
The final output of the layer is then given by
\begin{equation}
\mathbf{H}''=\mathbf{H}'+\operatorname{Concat}(\Delta\mathbf{H}_V^{f},\Delta\mathbf{H}_T^{f}),
\label{eq:terminate_ffn_residual}
\end{equation}
so that visual tokens remain unchanged throughout this layer, while textual tokens are updated as usual.

\subsection{RoPE: enabled or disabled?}
\label{appendix:rope}
% 为了探究在不同阶段启用与禁止位置编码对于模型性能的影响，我们在POPE和ChartQA数据集中对Qwen25-VL-7B模型中的各模块(IVR和SSR)中需要使用注意力分数的位置分别启用和禁用位置编码，得到表16的结果。结果显示，当我们在IVR模块禁用位置编码时可以提升模型性能，这说明在模型早期层存在位置偏差，其原因与位置编码相关，当我们禁用位置编码时可以在一定程度上缓解位置偏差。此外，当我们在SSR模块启用位置编码后可以获得更好的性能。这说明模型在第二阶段全局聚合后，注意力机制已经可以有效地捕捉正确的关键信息，其位置偏差在模型前向推理过程中减弱。上述实验结果启示我们可以根据模型在推理过程中的不同位置启用和禁用位置编码，以提高模型的定位、推理能力。
To study the impact of enabling or disabling positional encoding at different stages, we conduct experiments on \textbf{POPE} and \textbf{ChartQA} with Qwen2.5-VL-7B. Specifically, for both the IVR and SSR modules, we toggle positional encoding at the points where attention scores are used for decision making, and report the results in Table \ref{tab:rope}. The results show that disabling positional encoding in the IVR module improves performance, suggesting that early layers suffer from positional bias induced by positional encoding, and removing it can partially mitigate this bias. In contrast, enabling positional encoding in the SSR module yields better performance, indicating that after Stage-II global aggregation, the attention mechanism can more reliably capture the truly relevant information and positional bias becomes less pronounced during forward inference. These findings suggest that positional encoding should be configured in a stage-aware manner, enabling or disabling it at different points of inference to improve localization and reasoning.

\begin{table*}[]
    \centering
    \resizebox{\textwidth}{!}{
    \begin{tabular}{ccccccccccc}
    \toprule
    \multirow{2}{*}{\textbf{Setting}}&\multirow{2}{*}{\textbf{IVR}}&\multirow{2}{*}{\textbf{SSR}}&\multirow{2}{*}{\textbf{w/ rope}}&\multicolumn{4}{c}{\textbf{POPE}}&\multicolumn{3}{c}{\textbf{ChartQA}}\\
    \cmidrule(lr){5-8}
    \cmidrule(lr){9-11}
    & & & &Accuracy&F1 Score&Precision&Recall&Augmented Split&Human Split&Overall\\   \midrule
    $R_\text{IVR}=0.5,R_\mathcal{S}=0.2$&\CheckmarkBold&\XSolidBrush&\CheckmarkBold&86.94&85.50&96.14&76.98&92.64$\pm$0.74&64.56$\pm$1.35&78.60$\pm$0.82\\
    $R_\text{IVR}=0.5,R_\mathcal{S}=0.2$&\CheckmarkBold&\XSolidBrush&\XSolidBrush&87.12&85.71&96.24&77.27&92.16$\pm$0.76&65.12$\pm$1.35&78.64$\pm$0.82\\
    $R_\text{IVR}=0.25,R_\mathcal{S}=0.1$&\CheckmarkBold&\XSolidBrush&\CheckmarkBold&86.14&84.41&96.46&75.04&83.20$\pm$1.06&48.16$\pm$1.41&65.68$\pm$0.95\\
    $R_\text{IVR}=0.25,R_\mathcal{S}=0.1$&\CheckmarkBold&\XSolidBrush&\XSolidBrush&86.32&84.65&96.45&75.42&82.8$\pm$1.07&49.12$\pm$1.41&66.00$\pm$0.95\\
    $R_\text{SSR}=0.15$&\XSolidBrush&\CheckmarkBold&\CheckmarkBold&87.70&86.45&96.34&78.40&93.60$\pm$0.69&72.32$\pm$1.27&82.96$\pm$0.75\\
    $R_\text{SSR}=0.15$&\XSolidBrush&\CheckmarkBold&\XSolidBrush&87.63&86.36&96.26&78.31&93.76$\pm$0.68&71.76$\pm$1.27&82.76$\pm$0.76\\
    $R_\text{SSR}=0.05$&\XSolidBrush&\CheckmarkBold&\CheckmarkBold&87.73&86.48&96.32&78.47&92.96$\pm$0.72&70.00$\pm$1.30&81.48$\pm$0.78\\
    $R_\text{SSR}=0.05$&\XSolidBrush&\CheckmarkBold&\XSolidBrush&87.69&86.42&96.31&78.38&92.80$\pm$0.73&69.84$\pm$1.30&81.32$\pm$0.78\\
    % \XSolidBrush&\CheckmarkBold&\CheckmarkBold&
    \bottomrule
    \end{tabular}
    }
    \caption{Performance comparison of Qwen2.5-VL-7B under different settings when enabling or disabling RoPE in the IVR and SSR stages.}
    % 在不同设置下Qwen25-VL-7B的IVR和SSR阶段使用/不使用 Rope的性能对比
    \label{tab:rope}
\end{table*}

\section{Detailed Experiment Settings}
\label{sec:Experiment Settings}
\subsection{Datasets}
We performed experiments on eight widely used benchmarks, including GQA \cite{gqa}, MME \cite{mme}, POPE \cite{pope}, SQA \cite{sqa}, VizWiz \cite{vizwiz}, TextVQA ($\text{VQA}^{\text{text}}$) \cite{textvqa}, MMB \cite{mmb} and AI2D \cite{ai2d}. 

\noindent{\textbf{GQA}} consists of three key components: scene graphs, questions, and images. The image component includes the images themselves, as well as their spatial features and the attributes of all objects depicted within them. The questions in GQA are carefully designed to evaluate the model's ability to understand visual scenes and reason about various elements of the images.

\noindent{\textbf{MME}} is intended to rigorously assess a model's perceptual and cognitive capabilities through 14 distinct subtasks. It uses carefully designed instruction-answer pairs and clear instructions to minimize data leakage and ensure fair evaluation. This setup provides a reliable measure of a model's performance across a variety of tasks.

\noindent{\textbf{POPE}} is specifically designed to evaluate object hallucination. It consists of a series of binary questions regarding the presence of objects in images, using accuracy, recall, precision, and F1 score as metrics. This method provides a precise assessment of hallucination levels under various sampling strategies.

\noindent{\textbf{SQA}} covers a broad range of domains, including natural sciences, language, and social sciences. The questions are organized hierarchically into 26 topics, 127 categories, and 379 skills, offering a diverse and comprehensive framework for evaluating multimodal understanding, multi-step reasoning, and interoperability.

\noindent{\textbf{VizWiz}} is a visual benchmark created to support visually impaired individuals. It consists of real-world images taken by blind users, each paired with questions they ask about the images. The dataset includes 20,523 training, 4,319 validation, and 8,000 test image-question pairs, with each question having 10 human-annotated answers. VizWiz challenges models to either provide accurate answers to the questions or determine if a question is answerable, emphasizing practical visual understanding and accessibility.

\noindent{\textbf{TextVQA}} focuses on the integration of textual information within images. It tests a model's ability to read and reason about text embedded in visual content, requiring the model to understand both visual and textual elements in order to answer questions correctly.

\noindent{\textbf{MMB} provides a hierarchical evaluation framework that categorizes model capabilities into three levels. The first level (L-1) focuses on perception and reasoning. The second level (L-2) expands on this by introducing six sub-abilities, while the third level (L-3) further refines these into 20 specific dimensions. This structured approach enables a detailed and comprehensive assessment of a model’s diverse capabilities.

\noindent{\textbf{AI2D} is a dataset that contains over 5,000 elementary school science diagrams, with more than 150,000 rich annotations, their basic factual syntactic parsing, and over 15,000 corresponding multiple-choice questions.

\subsection{Models}
We evaluate HalfV using various open-source MLLMs. In particular, we assess our method on four prominent models: LLaVA-v1.5-7B\footnote{\url{https://huggingface.co/liuhaotian/llava-v1.5-7b}}, LLaVA-v1.5-13B\footnote{\url{https://huggingface.co/liuhaotian/llava-v1.5-13b}}  \cite{llava}, LLaVA-NeXT-7B\footnote{\url{https://huggingface.co/liuhaotian/llava-v1.6-mistral-7b}} \cite{llavanext}, and Qwen2.5-VL 7B-Instruct\footnote{\url{https://huggingface.co/Qwen/Qwen2.5-VL-7B-Instruct}} \cite{qwen25vl}.

\noindent{\textbf{LLaVA-v1.5}. LLaVA-v1.5 models process images with a 336$\times$336 resolution and treat each image as 576 tokens. Each image is divided into non-overlapping patches, with each patch being treated as a token, enabling the model to process the image in a way that aligns with the transformer architecture used for multimodal tasks.

\noindent{\textbf{LLaVA-NeXT}. LLaVA-NeXT divides high-resolution images into smaller subimages and encodes both the subimages and downsampled original images independently. This approach enables the model to scale the input to any arbitrary resolution without the need for positional embedding interpolation, which is typically required for Vision Transformers (ViTs) \cite{vit}. Compared to LLaVA-1.5, LLaVA-NeXT scales the input image resolution by 4× and increases the number of visual tokens by up to 5×, resulting in 2880 tokens per image.

\noindent{\textbf{Qwen2.5-VL}. Qwen2.5-VL's architecture is designed to handle large-scale vision-language tasks by efficiently processing high-resolution images, making it more capable of understanding and generating accurate descriptions of images. This makes it particularly effective for real-world applications that require both high visual fidelity and multimodal reasoning. Additionally, Qwen2.5-VL employs advanced attention mechanisms that allow it to efficiently scale to larger datasets while maintaining accuracy across a wide range of benchmarks.

\subsection{Baselines}
To evaluate the effectiveness of HalfV, we compare it with mainstream methods based on token-level redundancy and layer-level redundancy. Token-level redundancy-based methods mainly focus on how to select and prune visual tokens, which can be further divided into attention-based visual token pruning and diversity-based visual token pruning. In the comparative experiments, we use FastV \cite{fastv}, SparseVLM \cite{sparsevlm}, PDrop \cite{pdrop}, VisionZip \cite{visionzip}, HoloV \cite{holov}, and other attention-based methods, as well as DivPrune \cite{divprune} and other diversity-based visual methods as baselines. Additionally, we include layer-level redundancy-based visual token pruning methods such as BTP \cite{btp} and VTW \cite{vtw} for comparison. As a novel approach based on layer locking, ShortV \cite{shortv} is also included as one of our baselines for comparison.

\noindent{\textbf{FastV} concentrates on pruning tokens in the early stages by utilizing attention maps, thereby significantly reducing computational costs in the initial layers.

\noindent{\textbf{SparseVLM} assesses token importance through cross-modal attention and incorporates adaptive sparsity ratios, along with an innovative token recycling mechanism.

\noindent{\textbf{PDrop} employs a progressive token-dropping approach throughout the model stages, creating a pyramid-shaped token structure that optimizes both efficiency and performance.

\noindent{\textbf{VisionZip} determines token importance using attention in the encoder and clusters the remaining tokens based on key similarity. 

\noindent{\textbf{DivPrune}} addresses the token pruning challenge in LMMs by formulating it as a Max-Min Diversity Problem (MMDP), selecting a subset of tokens that maximize diversity.

\noindent{\textbf{BTP}} reduces the number of vision tokens by pruning in multiple stages, initially focusing on the global impact on subsequent layers and later emphasizing the preservation of local output consistency.

\noindent{\textbf{VTW}} drops all visual tokens after the K-th layer, enabling only text tokens to engage in the subsequent layers.

\noindent{\textbf{ShortV}} uses a novel metric, Layer Contribution (LC), to identify ineffective layers in Multimodal Large Language Models (MLLMs) and freezes visual token updates in these layers, significantly reducing computational costs while maintaining performance.

\subsection{Implementation Details}
All of our experiments are conducted on NVIDIA GeForce RTX 4080 SUPER (32G) GPU. The implementation was carried out in Python 3.10, utilizing PyTorch 2.1.2, and CUDA 13.0. All baseline settings follow the original paper.

\subsection{Hyperparameters}
In the real test environment, four hyperparameters need to be defined for the inference process: the number of IVR layer $L_\text{IVR}$, the retention ratio $R_\text{IVR}$, the anchor retention ratio $R_\mathcal{S}$, and the SSR starting layer $L_\text{SSR}$. To facilitate the reproduction of our experimental results, we list the parameter settings used for different models in Table \ref{tab:halfV params}. Note that $L_\text{IVR}$ and $L_\text{SSR}$ correspond to the starting layers of Stage II and Stage III, respectively. 
% 在真实测试环境中执行推理过程时，需要定义三个超参数：裁剪层数K、裁剪率R和锁定层N。为了便于复现我们的实验结果，我们在表1中列出了不同模型的参数设置。需要注意的是，裁剪层数K和锁定层N分别对应模型的第二阶段起始层和第三阶段起始层。各阶段的详细识别过程请参见Sec2和附录1。
\begin{table}[]
    \centering
    \resizebox{\columnwidth}{!}{
    \begin{tabular}{ccccc}
    \toprule
         Model&$L_\text{IVR}$&$R_\text{IVR}$&$R_\mathcal{S}$&$L_\text{SSR}$\\
         \midrule
         LLaVA-1.5v-7B (32 Layers) &3&50\%&0.2&15\\
         LLaVA-1.5v-13B (40 Layers)&3&50\%&0.2&15\\
         LLaVA-NeXT-7B (32 Layers)&2&50\%&0.1&16\\
         Qwen25-VL-7B (28 Layers)&2&[25\%,5\%]&0.1&21\\
    \bottomrule
    \end{tabular}
    }
    \caption{Hyperparameters for HalfV}
    \label{tab:halfV params}
\end{table}

%评测的其他baseline方法，具体实验参数如表2所示。对于单次视觉token裁剪方法，如FastV，FitPrune、Sparse VLM，需要指定裁剪层数K和裁剪后保留token的比例R。对于多次视觉token裁剪方法，需要指定裁剪层数列表及每次裁剪后保留的token数的比例。对于VTW方法的复现，我们按照原论文中的方法，选择在L/2处撤出V所有视觉token。其中L为总层数。对于ShortV的复现，我们按照原论文中所选择的层数N进行视觉token冻结。
The experiment also includes other baseline methods, with specific experimental parameters shown in Table \ref{tab:baseline setting}. For single-stage visual token pruning methods, such as FastV, FitPrune, and SparseVLM, the pruning layer number $K$ and the pruning ratio $R$. For multi-stage visual token pruning methods, the list of pruning layers and the ratio of retained tokens after each pruning stage must be provided. For the reproduction of the VTW method, we follow the approach in the original paper, removing all visual tokens at the $L/2$ layer, where $L$ is the total number of layers. For the reproduction of the ShortV method, we lock the visual tokens based on the layer number $N$ selected in the original paper.

% 在效率分析实验和模块效率的消融实验中，我们在LLaVA-NeXT-7B模型和Qwen25-VL-7B模型中均采用与主实验中相同的设置。在锚点保留率的敏感度实验中，我们在LLaVA-NeXT-7B模型中将总裁剪率设置为50\%，在Qwen25-VL-7B模型中将总裁剪率设置为77.8\%。在可视化的时候，我们令LLaVA-1.5v-7B和Qwen25-VL-7B模型中的总裁剪率均为77.8\%。在SSR起始层的敏感度实验中，我们令Qwen25-VL-7B保留稀疏token的比例为5\%。
In the efficiency evaluation and the module-efficiency ablation experiments, we use the same settings as in the main experiments for both LLaVA-NeXT-7B and Qwen2.5-VL-7B. For the anchor retention ratio sensitivity study, we set the overall pruning ratio to 50\% on LLaVA-NeXT-7B and 77.8\% on Qwen2.5-VL-7B. For visualization, we fix the overall pruning ratio to 77.8\% for both LLaVA-1.5v-7B and Qwen2.5-VL-7B. In addition, for the sensitivity study on the SSR starting layer, we set the sparse-token retention ratio to 5\% on Qwen2.5-VL-7B.
\begin{table}[]
    \centering
    \resizebox{\columnwidth}{!}{
    \begin{tabular}{ccccc}
    \toprule
    Model&Method&$R$&$K$&$N$\\
    \midrule
    \multirow{3}{*}{LLaVA-1.5v-7B}&Single-stage&33.3\%&2&-\\
    &Multi-stage&[50\%,75\%,87.5\%]&[5,10,20]&-\\
    &ShortV&-&-&19\\
    \midrule
    \multirow{3}{*}{LLaVA-1.5v-13B}&Single-stage&33.3\%&2&-\\
    &Multi-stage&[50\%,75\%,87.5\%]&[5,10,20]&-\\
    &ShortV&-&-&24\\
    \midrule
    \multirow{3}{*}{LLaVA-NeXT-7B}&Single-stage&33.3\%&2&-\\
    &Multi-stage&[50\%,75\%,87.5\%]&[5,10,20]&-\\
    &ShortV&-&-&24\\
    \midrule
    \multirow{3}{*}{Qwen25-VL-7B}&Single-stage&22.2\%&2&-\\
    &Multi-stage&[50\%,75\%,87.5\%]&[5,10,20]&-\\
    &ShortV&-&-&19\\
    \bottomrule
    \end{tabular}
    }
    \caption{Baseline Settings}
    \label{tab:baseline setting}
\end{table}

%在第三章的计算复杂度的计算中，需要用到文本token数t、视觉token数v。由于在真实测试环境中，文本token数是不固定的，在这里我们统一令t=50。此外，对于llava-1.5v系列模型，其生成的视觉token数固定为576。而对于llava-next和qwen系列的模型，其生成的视觉token数是动态变化的。我们根据真实场景中对于高分辨率图像的需求，将v设置为2352.其余参数的设置由各模型本身的属性决定，具体数据如表3所示。
\subsection{Computational Complexity}
\label{Computational Complexity}
We analyze the computational operations in the self-attention mechanisms and FFNs within the layers of the underlying LLM architecture. Let $t$ denote the count of text tokens, $v$ is the count of visual tokens, $h$ is the hidden state dimension, and $m$ is the intermediate dimension of the FFNs. The FLOPs in the three stages are calculated as:
\begin{equation}
\small
    \left\{
            \begin{array}{l}
                 F_{\text{I}}=2(t+v)(4h+3m)h+4(t+v)^2h, \\
                 \rule{0pt}{1.5em} F_{\text{II}}=2(t+v')(4h+3m)h+4(t+v')^2h, \\
                 \rule{0pt}{2.5em} F_{\text{III}}=\left\{
                \begin{aligned}
                & 2t(4h+3m)h + 4t(t+v')h,  \\
                & 2(t+v_{\text{ssr}})(4h+3m)h + 4(t+v_{\text{ssr}})^2h, 
                \end{aligned}
                \right.
            \end{array}
    \right.
\end{equation}
where $v'$ is the number of visual tokens after Stage II pruning. For Stage III, $F_{\text{III}}$ depends on the architecture-aware strategy: for \textit{Layer-level Inactivity} (e.g., LLaVA), visual updates are terminated ($v_{\text{active}}=0$), so only text tokens consume compute while attending to the full context $v'$; for \textit{Extreme Token Sparsity} (e.g., Qwen), computation is restricted to the top active visual tokens $v_{\text{ssr}}$ (where $v_{\text{ssr}} \ll v'$).
% In Stage I, the FLOPs of one dense Transformer layer can be calculated as:
% \begin{equation}
%     F_{\text{I}}=2(t+v)(4h+3m)h+4(t+v)^2h.
% \end{equation}
% Moving to Stage II, the FLOPs of one pruning layer, after reducing the number of visual tokens to $v'$, can be expressed as:
% \begin{equation}
%      F_{\text{II}}=2(t+v')(4h+3m)h+4(t+v')^2h.
% \end{equation}
% In Stage III, when visual tokens are locked, the FLOPs of one locking layer can be calculated as:
% \begin{equation}
%     F_{\text{III}}=2t(4h+3m)h+4t(t+v')h.
% \end{equation}
The total FLOPs for the entire process is then the weighted sum of the FLOPs for each stage, calculated as:
\begin{equation}
    FLOPs=L_{\text{I}}\times F_{\text{I}}+L_{\text{II}}\times F_{\text{II}}+L_{\text{III}}\times F_{\text{III}},
\end{equation}
where $L_{\text{I}}$, $L_{\text{II}}$, $L_{\text{II}}$ represent the number of layers for each corresponding stage.

In the computational complexity analysis of Section \ref{Computational Complexity}, the number of text tokens $t$ and visual tokens $v$ are required. Since the number of text tokens varies in real-world testing environments, we assume $t=50$ for consistency. Additionally, for the LLaVA-1.5v series models, the number of visual tokens generated is fixed at 576. However, for the LLaVA-NeXT and Qwen series models, the number of visual tokens is dynamically changing. Based on the demand for high-resolution images in real-world scenarios, we set $v=2352$. The other parameter settings are determined by the specific characteristics of each model, with the detailed data presented in Table \ref{tab:Computational settings}.

\begin{table}[]
    \centering
    \begin{tabular}{ccccc}
    \toprule
    Model&$t$&$v$&$h$&$m$\\
    \midrule
     LLaVA-1.5v-7B&50&576&4096&11008\\
     LLaVA-1.5v-13B&50&576&5120&13824\\
     LLaVA-NeXT-7B&50&2352&4096&14336\\
     Qwen25-VL-7B&50&2352&3584&18944\\
    \bottomrule
    \end{tabular}
    \caption{Computational Settings}
    \label{tab:Computational settings}
\end{table}
\section{Clarifications and Differences from Related Work}
Current approaches for accelerating multimodal large models during the prefill stage can be broadly categorized into two main directions: token-level acceleration and layer-level acceleration. The former studies how to design effective selection strategies that retain informative tokens while removing redundant ones to reduce computation. The latter focuses on identifying inefficient or redundant layers and lowering overall cost through layer skipping, early exiting, or freezing. We systematically analyze the strengths and limitations of both lines of work.

Token-level methods can significantly speed up prefill by retaining only a small number of visual tokens in early layers. However, under aggressive pruning ratios, especially for training-free schemes, model performance often degrades substantially, making the final outputs fall short of practical expectations. In contrast, layer-level methods typically preserve model quality more reliably, but their speedup is constrained by a clear bottleneck. When processing high-resolution images or videos that produce a large number of visual tokens, even keeping only a few dense layers can still incur considerable computation, so the overall acceleration often struggles to exceed 2×. Moreover, we find that such methods are highly model-dependent. For architectures with more active inter-layer representations and few ineffective layers, it is difficult to maintain original performance while sparsifying computation across layers.

Prior work usually designs acceleration strategies from a single perspective. Our observations suggest that a multi-level, model-aware acceleration design grounded in internal inference dynamics can improve efficiency while preserving performance to a much greater extent. In this section, we further clarify and distinguish our method from several related works that share partially similar designs from certain viewpoints.
\subsection{Comparison with ShortV}
% ShortV设计了一个层贡献指标，通过在预制数据集中计算模型各层的贡献度，从而确定冗余层。然后对于模型内部的冗余层，通过冻结视觉Token在冗余层的前向传递减少计算量。我们发现该方法严重依赖模型的内部冗余特性，对于存在较高冗余的LLaVA系列模型，该方法可以有效地通过避免视觉token在冗余层的矩阵计算，从而提高推理速度。但是当我们将该方法迁移至Qwen系列模型，即使仅冻结5-6层的相对冗余的层也会令模型的性能大幅下降。我们进一步对Qwen25-VL-7B模型各层间冗余度进行分析，发现Qwen模型内部冗余度比LLaVA模型低2-3个数量级。这说明Qwen系列模型其层间冗余度很低，因此基于层冗余的方法不适用此类模型。与之相反，我们发现对于Qwen这类低层冗余的模型也存在与LLaVA模型相同的视觉饱和现象，并且进一步发现该现象具体表现为极端Token稀疏性。我们显式利用该特性，实现在Qwen系列模型中的高加速比与高性能的平衡。
ShortV \citep{shortv} proposes a layer contribution metric to estimate how much each layer affects the model output on a calibration set, thereby identifying redundant layers. It then reduces computation by freezing visual tokens in these layers, avoiding their matrix operations during the forward pass. We find that this strategy is highly dependent on the model’s internal redundancy structure. For the LLaVA family, where inter-layer redundancy is relatively high, ShortV can effectively skip visual-token computation in redundant layers and thus improve inference speed.

However, when transferring this approach to the Qwen family, freezing even 5--6 relatively redundant layers leads to a substantial performance drop. We further analyze the layer-wise redundancy of Qwen2.5-VL-7B and observe that its redundancy is 2--3 orders of magnitude lower than that of LLaVA. This indicates that Qwen models exhibit very limited inter-layer redundancy, making layer-redundancy-based acceleration less suitable.

In contrast, we observe that Qwen, despite its low layer redundancy, still exhibits a visual saturation phenomenon similar to LLaVA. Moreover, in Qwen this phenomenon manifests as extreme token sparsity. We explicitly leverage this architecture-dependent property to achieve a better balance between high speedup and strong performance on the Qwen family.
\subsection{Comparison with BTP}
BTP \cite{btp} applies multi-stage pruning and selects pruning layers via inter-layer cosine similarity on a fixed sample set. It thus relies on a static, data-driven layer profile. By contrast, our method is grounded in the functional lifecycle of redundancy. We explicitly align acceleration with how redundancy emerges and evolves. 
\subsection{Comparision with PDrop}
% PDrop 的主要贡献在于揭示了一个关键现象：模型对浅层的视觉 token 裁剪更为敏感，而在更深层进行裁剪则相对更具鲁棒性。然而，该工作并未对这一现象给出更深入的机制性解释。相比之下，我们利用截断矩阵熵（TME）对其成因进行了剖析，指出其根本原因在于模型内部存在一个全局聚合阶段；在该阶段之后，关键信息能够被集中到少量视觉 token 中，从而使深层裁剪更不容易破坏模型的有效表征。
% 此外，PDrop 采用均匀分段的方式，基于直觉在从浅到深的过程中逐级裁剪 token。与之不同的是，我们的方法能够依据模型的内部特性，自动定位更合适的处理层，从而实现更具针对性的裁剪策略。实验结果也表明，在相近的加速比下，我们的方法能够更好地保持模型性能，这一优势来源于我们以模型内部机制为依据的设计，而非仅凭经验直觉进行分段裁剪。
PDrop’s \cite{pdrop} key contribution is to reveal an important phenomenon: LVLMs are much more sensitive to visual-token pruning in shallow layers, while pruning in deeper layers is considerably more robust. However, PDrop does not further investigate the underlying mechanism behind this observation. In contrast, our work provides a principled explanation through Truncated Matrix Entropy (TME). We show that the robustness in deep layers stems from a global aggregation stage inside the model, after which critical information can be consolidated into a small subset of visual tokens.

Moreover, PDrop adopts a uniform stage partition and prunes tokens progressively from shallow to deep layers based on intuition. Different from this heuristic design, our method leverages internal model properties to identify more appropriate processing layers. Empirically, we achieve better performance than PDrop under comparable speedup ratios, which we attribute to our model-aware design rather than an intuitive, uniform pruning schedule.
\section{Ethics Statement}
This work utilizes public datasets and open-source models, and we identify no significant ethical issues regarding data privacy or potential misuse. In accordance with the ACL Policy on AI Writing Assistance, we declare that AI tools were employed solely for the purpose of refining the writing style, correcting grammatical errors, and improving readability. The generation of core ideas, experimental design, data analysis, and the formulation of scientific conclusions were conducted entirely by the human authors.

\end{document}